%% file: Main.tex
\documentclass[sigconf, nonacm, pdfa]{acmart}
\newcommand\vldbdoi{XX.XX/XXX.XX}
\newcommand\vldbpages{XXX-XXX}
\newcommand\vldbvolume{17}
\newcommand\vldbissue{11}
\newcommand\vldbyear{2024}
\newcommand\vldbauthors{\authors}
\newcommand\vldbtitle{\shorttitle} 
\newcommand\vldbavailabilityurl{https://github.com/K-Coconut/MCPBenchmark}
\newcommand\vldbpagestyle{empty}

\usepackage{subcaption}
\usepackage{booktabs}
\usepackage{tablefootnote}
\usepackage{algorithm}
\usepackage{algpseudocode}
\usepackage{amsmath}
\usepackage{fancyhdr}
\usepackage{natbib}
\usepackage[show]{notebox}
\usepackage{arydshln}
\usepackage{color}
\usepackage{float}
\usepackage{balance}
\usepackage{enumitem}
\setlist{nosep}

\newtheorem{problem}{Problem}

\begin{document}
\title{A Benchmark Study of Deep-RL Methods for Maximum Coverage Problems over Graphs}


\author{Zhicheng Liang}
\email{zhicliang3-c@my.cityu.edu.hk}
\orcid{0009-0008-9015-8907}
\affiliation{
  \institution{City University of Hong Kong}
  \city{Hong Kong}
  \country{China}
}

\author{Yu Yang}
\email{yuyang@cityu.edu.hk}
\orcid{0000-0002-8209-2898}
\affiliation{
  \institution{City University of Hong Kong}
  \city{Hong Kong}
  \country{China}
}

\author{Xiangyu Ke}
\email{xiangyu.ke@zju.edu.cn}
\orcid{0000-0001-8082-7398}
\affiliation{
  \institution{Zhejiang University}
  \city{Hangzhou}
  \country{China}
}

\author{Xiaokui Xiao}
\email{xkxiao@nus.edu.sg}
\orcid{0000-0003-0914-4580}
\affiliation{
  \institution{National University of Singapore}
  \city{Singapore}
  \country{Singapore}
}

\author{Yunjun Gao}
\email{gaoyj@zju.edu.cn}
\orcid{0000-0003-3816-8450}
\affiliation{
  \institution{Zhejiang University}
  \city{Hangzhou}
  \country{China}
}



\input{abstract}

\maketitle

\pagestyle{\vldbpagestyle}
\begingroup\small\noindent\raggedright\textbf{PVLDB Reference Format:}\\
\vldbauthors. \vldbtitle. PVLDB, \vldbvolume(\vldbissue): \vldbpages, \vldbyear.\\
\href{https://doi.org/\vldbdoi}{doi:\vldbdoi}
\endgroup
\begingroup
\renewcommand\thefootnote{}\footnote{\noindent
Z. Liang and Y. Yang contribute equally. X. Ke is the corresponding author. This work was done while X. Ke worked at the National University of Singapore. \\ 
}
\renewcommand\thefootnote{}\footnote{\noindent
This work is licensed under the Creative Commons BY-NC-ND 4.0 International License. Visit \url{https://creativecommons.org/licenses/by-nc-nd/4.0/} to view a copy of this license. For any use beyond those covered by this license, obtain permission by emailing \href{mailto:info@vldb.org}{info@vldb.org}. Copyright is held by the owner/author(s). Publication rights licensed to the VLDB Endowment. \\
\raggedright Proceedings of the VLDB Endowment, Vol. \vldbvolume, No. \vldbissue\ %
ISSN 2150-8097. \\
\href{https://doi.org/\vldbdoi}{doi:\vldbdoi} \\
}\addtocounter{footnote}{-1}\endgroup

\ifdefempty{\vldbavailabilityurl}{}{
\begingroup\small\noindent\raggedright\textbf{PVLDB Artifact Availability:}\\
The source code, data, and/or other artifacts have been made available at \url{\vldbavailabilityurl}.
\endgroup
}

\input{1-Intro}
\input{2-Pre}
\input{3-Algo}

\input{4a-Exp_Basic.tex}
\input{4b-Exp_Limit.tex}
\input{5-Related}

\input{6-Con.tex}

\begin{acks}
This work was supported in part by the Hong Kong Research Grants Council (ECS 21214720), Alibaba Group through Alibaba Innovative Research (AIR) Program, Zhejiang Province's ``Lingyan'' R\&D Project under Grant No. 2024C01259, the Ningbo Yongjiang Talent Introduction Programme (2022A-237-G), the Ministry of Education, Singapore, under its MOE AcRF TIER 3 Grant (MOE-MOET32022-0001). Any opinions, findings, and conclusions or recommendations expressed in this material are those of the author(s) and do not reflect the views of the funding agencies.
\end{acks}

\bibliography{Main}

\input{7-Appendix.tex}


\end{document}

%% file: abstract.tex
\begin{abstract}
Recent years have witnessed a growing trend toward employing deep reinforcement learning (Deep-RL) to derive heuristics for combinatorial optimization (CO) problems on graphs. Maximum Coverage Problem (MCP) and its probabilistic variant on social networks, Influence Maximization (IM), have been particularly prominent in this line of research. In this paper, we present a comprehensive benchmark study that thoroughly investigates the effectiveness and efficiency of five recent Deep-RL methods for MCP and IM. These methods were published in top data science venues, namely S2V-DQN, Geometric-QN, GCOMB, RL4IM, and LeNSE. 
Our findings reveal that, across various scenarios, the Lazy Greedy algorithm {\em consistently outperforms all} Deep-RL methods for MCP. In the case of IM, theoretically sound algorithms like IMM and OPIM demonstrate superior performance compared to Deep-RL methods in most scenarios. Notably, we observe an {\em abnormal phenomenon} in IM problem where Deep-RL methods slightly outperform IMM and OPIM when the influence spread nearly does not increase as the budget increases.
Furthermore, our experimental results highlight common issues when applying Deep-RL methods to MCP and IM in practical settings. Finally, we discuss potential avenues for improving Deep-RL methods.
Our benchmark study sheds light on potential challenges in current deep reinforcement learning research for solving combinatorial optimization problems.


\end{abstract}

%% file: 1-Intro.tex
\section{Introduction}
\label{sec:intro}

Combinatorial optimization problems play a pivotal role in addressing complex challenges across diverse domains. The {\em Maximum Coverage Problem} (MCP) and its probabilistic variant, {\em Influence Maximization} (IM), stand out as representative challenges in this realm.
The primary goal of MCP is to {\em identify a set $S$ of $k$ nodes from a given input graph $G$ to maximize node coverage}. On the other hand, IM focuses on maximizing influence spread in social networks by selecting a strategic set of nodes. 
The significance of MCP and IM transcends theoretical domains, finding practical applications in scheduling \cite{guihaire2008transit, erdougan2010scheduling}, opinion formation and election ~\cite{ke2023opinion}, facility location \cite{chauhan2019maximum, boonmee2017facility}, recommendation systems \cite{habibi2015keyword, chen2015information,yang2019tracking}, viral marketing~\cite{kempe2003maximizing, yang2020continuous,yang2019influence, ke2018influence, Ke24Host}, and sensor placement~\cite{leskovec2007cost, ke2022reliability}. 
Therefore, devising effective and efficient algorithms for solving MCP and IM has drawn great attention from the data management and data mining research communities.

Existing studies have made creditable strides in understanding the algorithmic complexities of MCP and IM. 
A simple greedy algorithm can return a solution of $(1 - \frac{1}{e})$-approximation guarantee and Feige \cite{feige1998threshold} had demonstrated that achieving a better approximation ratio than $1 - \frac{1}{e}$ is unlikely unless $\textbf{P}=\textbf{NP}$. 
The greedy algorithm's applicability extends to IM, as illustrated by~\cite{kempe2003maximizing}. Borgs et al.~\cite{borgs2014maximizing} proposed a {\em Reverse Influence Sampling} (RIS) method that can achieve $1-\frac{1}{e}-\epsilon$ approximation for IM when equipped with the greedy search. RIS algorithm has a time complexity of $O((m+n)\log{n}/\epsilon^2)$, which is nearly optimal (up to a logarithmic factor) with respect to network size. Tang et al. further enhanced the practical efficiency of the RIS-based algorithms~\cite{tang2014influence,tang2015influence}. Tang et al. utilized the online approximation bound of submodular functions~\cite{tang2018online} to return online approximation bounds of RIS-based algorithms. As with MCP, any approximation ratio better than $1-\frac{1}{e}$ is implausible unless $\textbf{P}\neq \textbf{NP}$.

Despite the substantial theoretical advancements in MCP and IM, recently, there has been a notable surge in the application of Deep-RL to these combinatorial optimization problems~\cite{khalil2017learning,chen2021contingency,kamarthi2020influence,manchanda2020gcomb,he2025exploring,yang2025behaviour,ireland2022lense}. This contemporary approach seeks to harness the power of data-driven learning, aiming to derive heuristics that can surpass the performance of theoretically grounded algorithms in practice. In these studies, {\em Graph Neural Networks} (GNNs) initially learn node embeddings, followed by the integration of reinforcement learning components like Q-learning to approximate the objective function. 
Empirical studies of these works reveal instances where they outperform traditional algorithms for both MCP and IM.

\begin{figure}[t]\centering
    \vspace{-0.5mm}
    \begin{subfigure}[b]{0.48\linewidth}
        \includegraphics[width=\linewidth,height=0.95\linewidth]{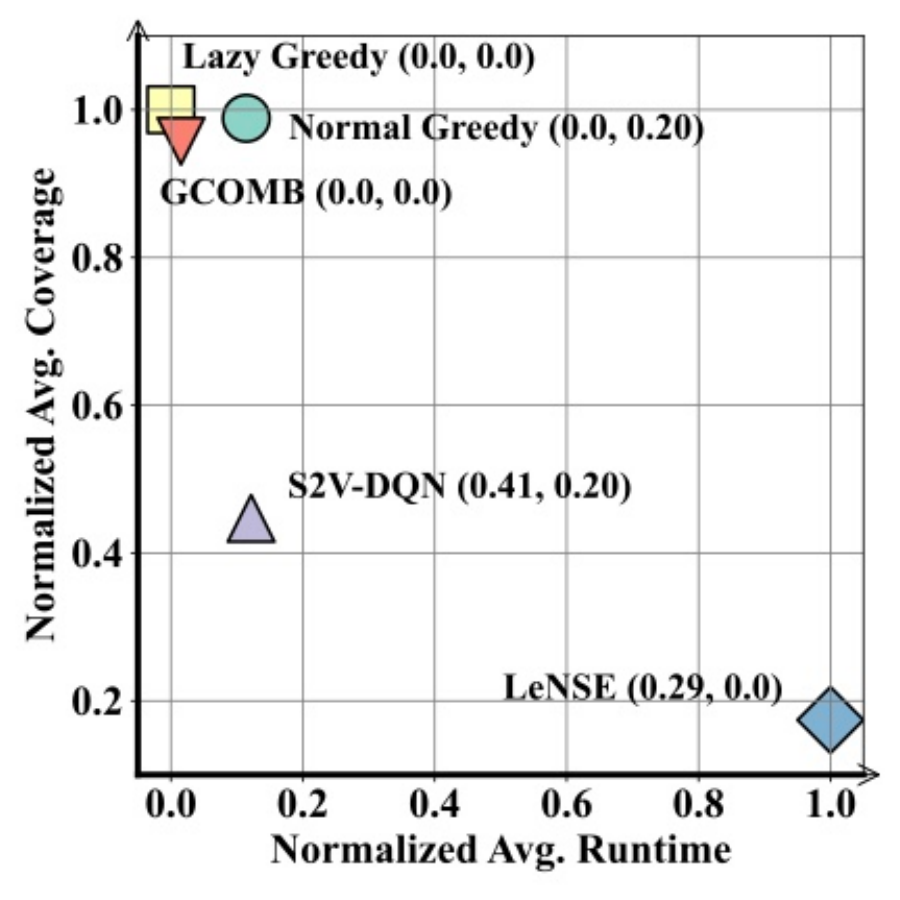}
        \vspace{-6mm}
        \caption{MCP}
    \end{subfigure}
    \begin{subfigure}[b]{0.48\linewidth}
        \includegraphics[width=\linewidth,height=0.95\linewidth]{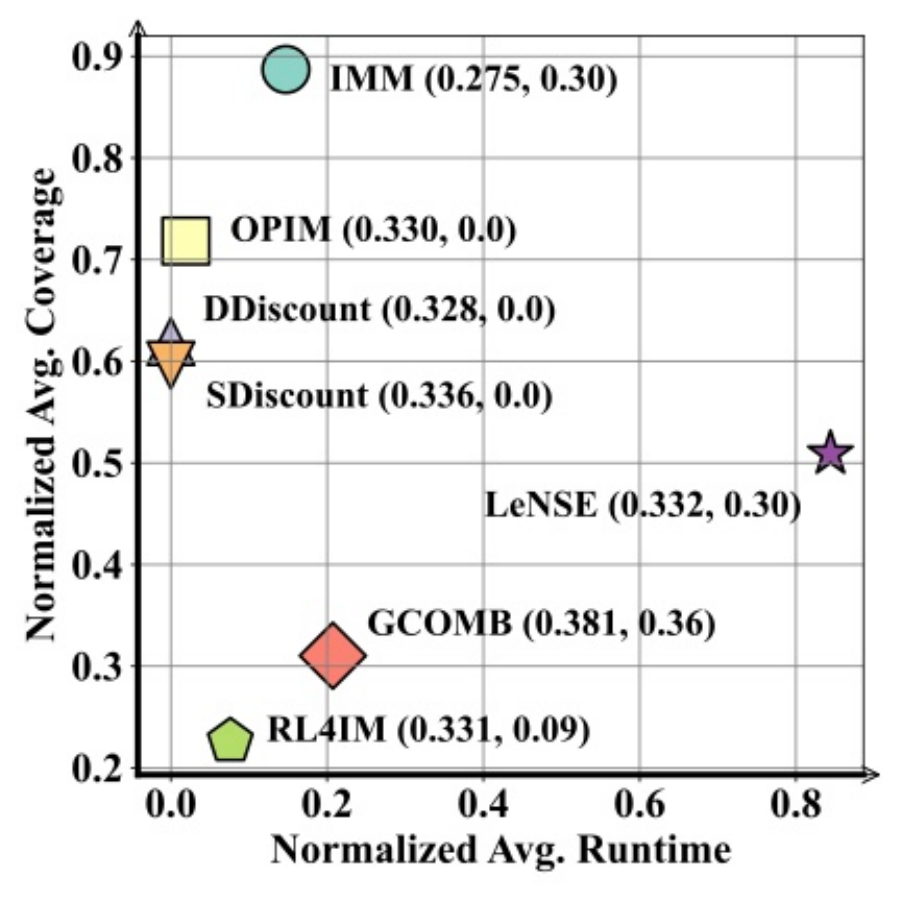}
        \vspace{-6mm}
        \caption{IM}
    \end{subfigure} 
    \vspace{-4mm}
    \caption{Performance overview of methods on MCP and IM: Values represented as \textbf{Method} (Coverage standard deviation, Runtime standard deviation). Geometry-QN was excluded from (b) due to the scalability issues on most graphs.}
    \vspace{-6mm}
\label{fig:overview_performance}
\end{figure}

An aspect of concern in the application of Deep-RL lies in the treatment of training time. Notably, the prevailing practice of excluding training time from computation cost evaluations raises valid questions. Viewing {\em training as akin to pre-processing}, it is customary in rigorous data management research to consider the pre-processing time in an amortized manner {\em within query computation cost assessments} \cite{malicevic2017everything}. 
The exclusion of training time might be justifiable only in scenarios where training is performed on a dataset of size independent of subsequent queries, rendering the training time a theoretically constant factor \cite{alon1987optimal, cormen2022introduction}. However, even in such cases, the impact of constant-time training or pre-processing might be limited in boosting the efficiency of a broader range of queries \cite{kearns1994introduction}. 
Consequently, there arises a legitimate concern about the {\em actual effectiveness and efficiency} of these Deep-RL methods, prompting a closer examination of their claimed benefits. 

\begin{figure*}[h]
  \centering  \includegraphics[width=.8\linewidth]{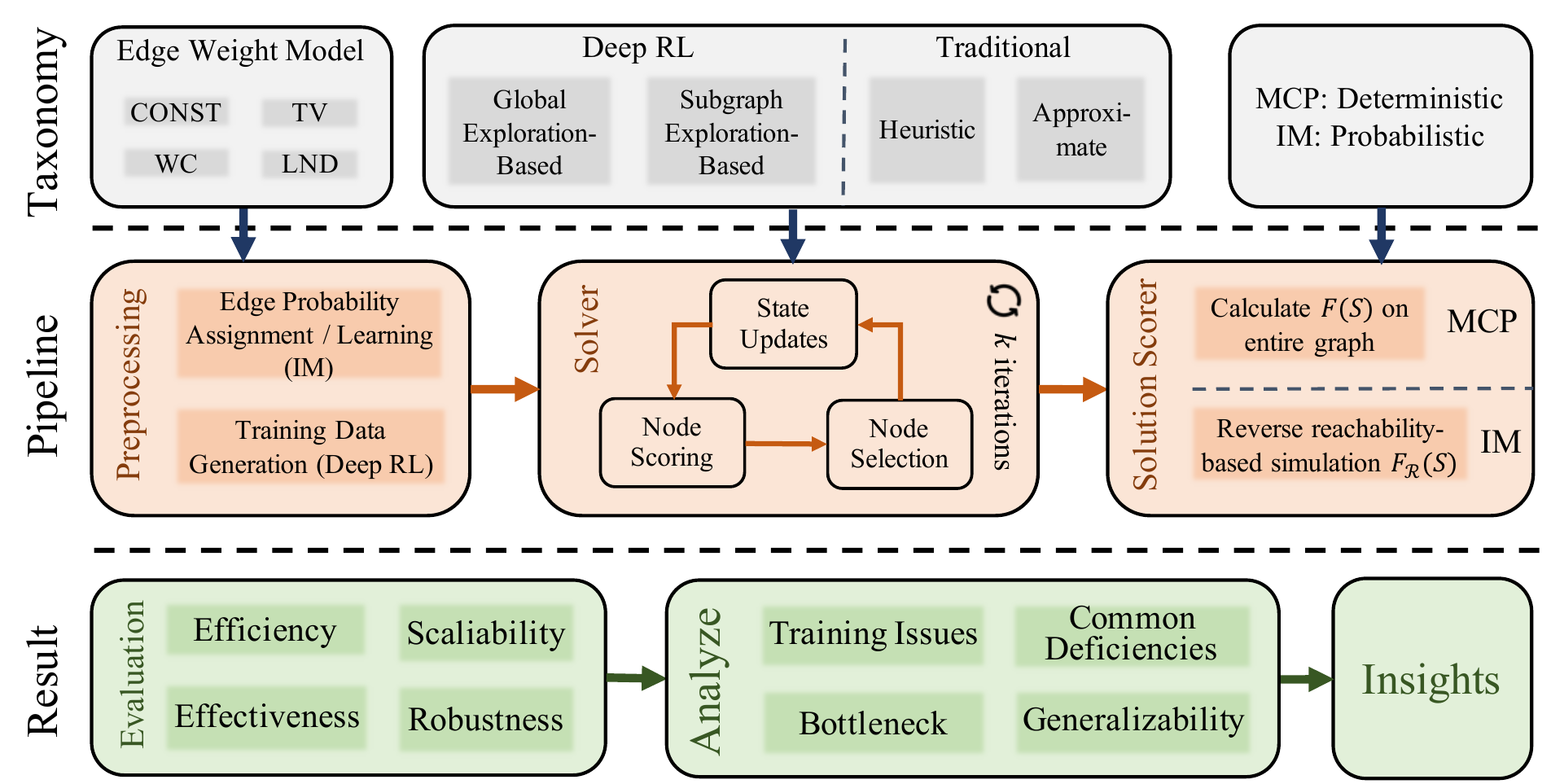}
  \vspace{-1mm}
  \caption{Benchmarking Framework}
  \vspace{-2.5mm}
  \label{framework}
\end{figure*}

To this end, this paper conducts a thorough benchmark study on recent Deep-RL methods~\cite{khalil2017learning,chen2021contingency,kamarthi2020influence,manchanda2020gcomb,ireland2022lense}. Fig.~\ref{framework} presents a comprehensive overview of the benchmarking framework. 
The pre-processing phase initiates the generation of training data as needed, and propagation probabilities are assigned as edge weights in IM based on diverse edge weight models \cite{kempe2003maximizing, arora2017debunking, manchanda2020gcomb}. 
Deep-RL solvers, categorized into global and subgraph exploration types, undergo training and validation before their applications. In tandem, traditional solvers encompass both approximate and heuristic approaches.
Subsequently, a solution scorer is deployed to calculate coverage. In the case of IM, the scorer estimates the influence spread $F_\mathcal{R}(S)$ based on {\sf RIS}-based simulations. Meanwhile, for MCP, the coverage $F(S)$ is directly calculated on the input graph. 
We evaluate on 20 commonly-used datasets and test 4 edge weight models.
Finally, a thorough analysis of the results is conducted from diverse perspectives, enabling a nuanced understanding of the strengths and limitations of Deep-RL methods compared to traditional solvers.

The contributions of this paper are summarized as follows: 

(1) We summarize the general pipeline of recent Deep-RL methods~\cite{khalil2017learning,chen2021contingency,kamarthi2020influence,manchanda2020gcomb,ireland2022lense} for MCP and IM in \S~\ref{sec:alg}. Additionally, in \S~\ref{sec:concern} and \ref{sec:lazy}, we articulate concerns regarding the exclusion of training time and the absence of a strong baseline, Lazy Greedy, for MCP in the Deep-RL studies, which motivate our study.

(2) \S~\ref{sec:benchmark} details an extensive benchmark study aimed at examining recent Deep-RL methods for MCP and IM. We successfully reproduce some of the experimental results reported in prior studies~\cite{khalil2017learning,chen2021contingency,kamarthi2020influence,manchanda2020gcomb,ireland2022lense} and expand more. Our findings, illustrated in Fig.~\ref{fig:overview_performance}, reveal that, despite reported successes, traditional and well-established MCP and IM algorithms often outperform these Deep-RL methods in terms of both effectiveness and efficiency.

(3) \S~\ref{sec:deeper} delves deeper into the practical challenges faced by Deep-RL methods. We highlight common issues, such as the difficulty in determining the suitability of a testing graph for a trained Deep-RL model and the observed performance fluctuations concerning training time and data size.

(4) Our experimental findings prompt a discussion on the challenges that must be addressed for Deep-RL methods to provide effective and efficient solutions for MCP and IM, offering insights for future research in this domain (\S~\ref{sec:solutions}).

For conciseness, certain details are omitted from the main text. 
These include the algorithms discussed, additional variants of MCP, certain evaluations such as those for GCOMB's noise predictor, strategies to enhance LeSNE's efficiency, and outcomes from more datasets. Readers are encouraged to refer to the appendices in our full version~\cite{LYKXG24Code}.

%% file: 2-Pre.tex
\vspace{-1mm}
\section{Preliminaries}
\label{section: preliminaries}
\vspace{-0.8mm}

In this section, we formally revisits the {\em Maximum Coverage Problem} (MCP) and its probabilistic variant {\em Influence Maximization} (IM).



\vspace{-1mm}
\subsection{Problem Statements}
\vspace{-0.8mm}

\begin{problem}[Maximum Coverage Problem (MCP) on Graphs]
Given a graph $G = (V,E)$, let $f(S)=\frac{|X_S|}{|V|}$ be the coverage function where $X_S=\{j|j \in S \vee \exists (i,j)\in E,i\in S\}$. For a given budget $k$, we aim at selecting a set $S \subseteq V$, $|S| = k$ to maximize the coverage $f(S)$.
\end{problem}
\vspace{-1mm}

The IM problem can be viewed as a maximum coverage or reachability problem on a probabilistic graph~\cite{XAL19}. IM primarily revolves around modeling the dynamics of influence diffusion within a network. In this particular paper, our focus is on the Independent Cascade IC model \cite{kempe2003maximizing}, which is widely recognized as the most popular influence diffusion model in the literature.

\noindent \textbf{Independent Cascade (IC) Model.} Given a weighted and directed graph $G=\langle V, E, W \rangle$, we assume that the edge weight $p_{uv}$ on an edge $(u,v)$ represents its \emph{influence probability} $p_{uv}$, i.e., $0 \leq p_{uv}=w_{uv} \leq 1$. An influence diffusion starts from a \textbf{seed set} $S \subseteq V$. $S_i$ denotes the set of nodes that are active in time-step $i$, $i\in \mathbb{N}$, $S_0=S$. Each newly-activated vertex $u$ in the previous step $i-1$ has a single chance at the current step $i$ to influence its inactive out-neighbor $v$ independently with a probability $p_{uv}$. The diffusion process continues until there are no further activations, i.e., $S_t = S_{t-1}$. The \textbf{influence spread} of $S$, $I(S)$, is the expected number of active nodes at the end of the diffusion initiated from $S$. As IC model is the most popular influence model, we employ IC model to formulate the IM problem.

\vspace{-1.8mm}
\begin{problem}[Influence Maximization (IM)] Given a social network $G=(V, E, W)$, a budget $k$, we want to select a set $S \subseteq V$, $|S| = k$ such that the influence spread $I(S)$ under IC model is maximized.
\end{problem}
\vspace{-1mm}

\vspace{-2mm}
\subsection{Characteristics of MCP and IM}
\vspace{-0.8mm}

\noindent \textbf{Intractability.} Both MCP and IM problems are known to be {\bf NP}-hard~\cite{kempe2003maximizing}. An essential characteristic shared by both is the presence of {\em monotonicity} and {\em submodularity} in their objective functions~\cite{kempe2003maximizing}. These properties enable the development of efficient approximation algorithms efficient approximation algorithms, guaranteeing a performance ratio of $1-\frac{1}{e}$~\cite{kempe2003maximizing, tang2015influence}.
The established understanding is that achieving an approximation ratio higher than $1-\frac{1}{e}$ is implausible unless the complexity classes {\bf P} and {\bf NP} are equal~\cite{kempe2003maximizing}.

\noindent \textbf{Influence Spread Estimation.} In addition to the {\bf NP}-hardness, computing the influence spread under the IC model for IM problem is known to be {\bf \#P}-hard~\cite{chen2010scalable}. To address this challenge, the polling method~\cite{borgs2014maximizing} emerges as an efficient solution, ensuring a $1-\frac{1}{e}-\epsilon$ approximation guarantee for IM with high probability. In the Polling method, the estimation of influence spread involves the sampling of {\sf Reverse Reachable} (RR) sets~\cite{tang2014influence}, outlined as follows. 
Each edge $(u,v)$ in the graph is independently preserved with a probability $p_{uv}$. The RR set for vertex $v$ consists of all nodes that can reach $v$ through the preserved edges.
To estimate the influence spread, the polling method involves sampling a specified number, denoted as $M$, of random RR sets. Each RR set is constructed as described above. The quantity $D(S)$ represents the number of RR sets that contain at least one vertex from a given set $S$. It has been shown that $\frac{|V|D(S)}{M}$ provides an unbiased estimation of the influence spread $I(S)$~\cite{borgs2014maximizing, tang2014influence}.
In practical applications, a large sample size $M$ ensures an accurate estimation of the influence spread $I(S)$ with a high probability. 

\vspace{-3mm}
\subsection{Edge Weight Models in IM}

The IM problem involves a weighted graph where edge weights $p_{uv}$ denote the direct influence from node $u$ to node $v$. However, acquiring accurate data for learning these edge weights is often challenging. In the IM literature, predefined models are commonly employed to address this issue:

\noindent \textbf{Tri-valency (TV) Model~\cite{manchanda2020gcomb}.} The weight of an edge is chosen randomly from a set of weights \{0.001, 0.01, 0.1\}. 

\noindent \textbf{Constant (CONST) Model~\cite{manchanda2020gcomb}.} In this model, The weight of an edge is set as a constant value, e.g., 0.1. 

\noindent \textbf{Weighted Cascade (WC) Model~\cite{kempe2003maximizing}.} The weight of the edge (u, v) is set as $\frac{1}{|N^{in}(v)|}$, where $N^{in}(v)$ is the set of $v$'s in-neighbors. 

\noindent \textbf{Learned (LND) Model~\cite{manchanda2020gcomb}.} When we have historical data of user interactions, we learn edge weights. One representative method to learn the weights is the Credit Distribution Model~\cite{goyal2011data}.

\noindent \textbf{Remark.} It's noteworthy that for theoretically sound algorithms~\cite{tang2014influence,tang2015influence,tang2018online}, {\em the specific choice of edge weight setting doesn't impact the effectiveness of the algorithms}. However, our benchmark study reveals that the choice of edge weight setting can influence the performance of Deep-RL heuristics \S \ref{sec:performance_im}.

%% file: 3-Algo.tex
\vspace{-1mm}
\section{Algorithms Revisited}\label{sec:alg}
\vspace{-0.5mm}

\begin{figure*}[h]
  \centering
  \vspace{-1mm}
  \includegraphics[width=.95\linewidth]{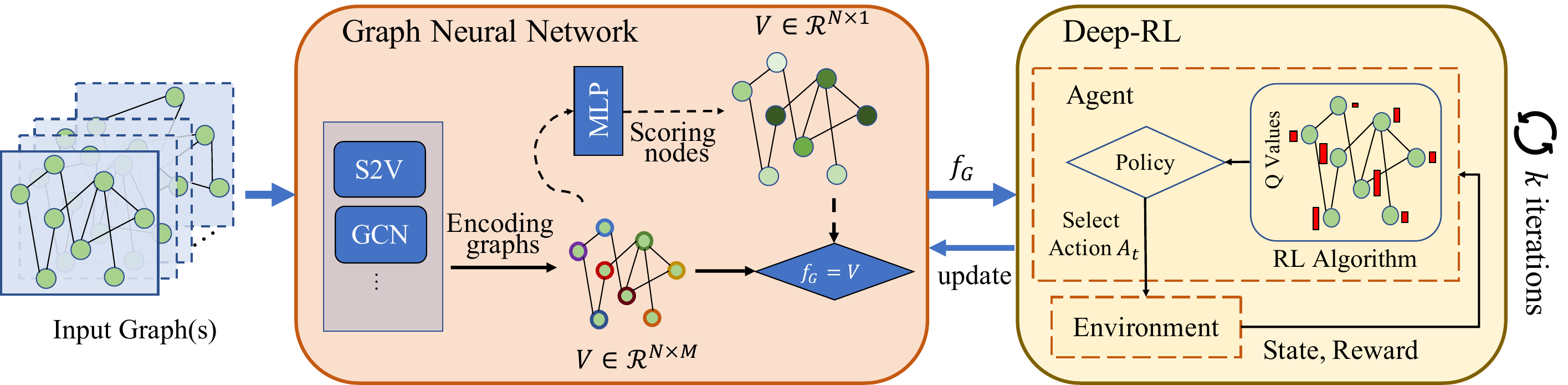}
  \caption{General pipeline of Deep-RL methods for solving MCP/IM over Graphs}
  \label{pipeline}
  \vspace{-1mm}
\end{figure*}

In this section, we delve into the foundational concepts of Deep-RL and reflect upon the Deep-RL techniques and other traditional algorithms used for the MCP/IM benchmarks in our experiments, and we also raise concerns on these studies.

\vspace{-1mm}
\subsection{Deep-RL with GNNs}
\vspace{-0.5mm}

\noindent \textbf{GNNs.} Graph Neural Networks (GNNs)~\cite{liao2021review} are designed to process data in graph structures, capturing relationships and features of nodes and edges. Unlike traditional neural networks, GNNs handle irregular structures, making them ideal for graph-structured data like social networks. By propagating information through the graph, GNNs capture both local and global structures.

\noindent \textbf{Deep-RL.} Reinforcement Learning (RL) \cite{vanOtterlo2012} is a prominent branch of machine learning where agents learn optimal strategies through interactions with an environment, receiving feedback in the form of rewards or penalties. The fundamental goal of RL is to deduce a policy that maximizes the expected cumulative reward over time, emphasizing decision-making to achieve specific objectives. 
In this dynamic process, an agent, situated in a given environmental state, takes actions and receives rewards based on the outcomes, commonly modeled as a Markov Decision Process (MDP). However, dealing with intricate or high-dimensional environments poses challenges. 
To address this, Deep-RL integrates deep neural networks, enabling the extraction of features from raw data and the approximation of complex functions. This fusion enhances RL's capability to tackle large-scale, high-dimensional problems.

\subsection{Deep-RL Methods for MCP/IM}

We outline a systematic approach for employing Deep-RL methods to address coverage problems on graphs, illustrated in Fig.~\ref{pipeline}. 
The procedure initiates with a GNN encoder, tasked with learning the graph's embedding. Subsequently, an MLP layer utilizes this embedding to compute scores for each node, represented as either the node embedding matrix or the vector of node scores, denoted as $f_G$. 
Using the generated node scores, a series of seed sets are randomly created, serving as training data for the RL model. In the RL environment, the feature $f_G$ defines the state, facilitating reward computation. The RL agent, guided by a learned policy, iteratively seeks the optimal solution set $S$ that maximizes the overall reward, with a cardinality of $|S|=k$. Subsequent paragraphs elaborate on how the methods discussed in this paper tailor this general pipeline to address specific aspects of the coverage problem on graphs.

\noindent \textbf{S2V-DQN}~\cite{khalil2017learning}. It first maps the input graph $G$ into node embeddings via Struc2Vec \cite{dai2016discriminative}. A deep Q-network (DQN)~\cite{mnih2013playing, mnih2015human}  is learned to construct a solution set that maximizes the coverage based on the node embedding.

\noindent \textbf{RL4IM}~\cite{chen2021contingency}. Unlike S2V-DQN, RL4IM utilizes Struc2Vec to encode graph-level features rather than node-level ones. The input graph $G$ is randomly selected from a set of training graphs $\mathcal{G}$ in each iteration. The reward for each action is calculated by Monte Carlo (MC) simulations on the fly. Two novel tricks, namely state abstraction and reward shaping, are used to improve performance.


\vspace{0.3mm}
\noindent \textbf{Geometric-QN}~\cite{kamarthi2020influence}. Starting from a randomly initialized seed set $S$, Geometric-QN iteratively enlarges a subgraph $g$ by a random walk over the input graph $G$. Then DeepWalk~\cite{perozzi2014deepwalk} is used to generate node features and a GCN~\cite{kipf2017semisupervised} encoder excavates the structural information. Lastly, a DQN selects the node with the highest Q value and adds it into $S$, making it possible to expand $g$ to cover more potentially influential nodes.

\vspace{0.3mm}
\noindent \textbf{GCOMB}~\cite{manchanda2020gcomb}. Rather than unsupervised learning, GCOMB trains a GCN network~\cite{kipf2017semisupervised} in a supervised manner, where the label of each node is generated by calculating its marginal cover (influence spread) based on a probabilistic greedy method. A DQN then finds a solution set to maximize the cover (influence spread). Node pruning techniques are also adopted to remove noisy nodes to reduce the search space, which makes GCOMB scalable to large-scale graphs.

\vspace{0.3mm}
\noindent \textbf{LeNSE} \cite{ireland2022lense}. Similar to Geometry-DQN, LeNSE aims to find a smaller optimal subgraph containing the optimal solution set. It generates multiple subgraphs with a fixed number of nodes, categorizes them into labels based on the likelihood of containing the optimal solution, and trains a GNN to cluster similar subgraphs. By leveraging both node-level and graph-level features generated by GNN, a DQN constructs the subgraph iteratively. Finally, a pre-existing heuristic is applied to discover a solution set from the constructed subgraph.

\vspace{-2mm}
\subsection{Traditional Algorithms for MCP/IM}
\vspace{-0.8mm}

\noindent \textbf{Greedy for MCP}. \label{sec: lazy}  Greedy algorithm~\cite{kempe2003maximizing} sequentially selects a node that covers the most remaining uncovered elements. Leveraging the submodularity of coverage functions, {\sf Lazy Greedy} (also known as {\sf CELF}~\cite{leskovec2007cost}) distinguishes itself by strategically minimizing computational overhead. After initial marginal gain computations for all nodes, it selectively updates and reevaluates only the top contenders in subsequent iterations. This approach retains the $(1-\frac{1}{e})$ approximation guarantee while delivering a significant speedup over the traditional greedy algorithm. (More details can be found in Appendix of our full paper~\cite{LYKXG24Code}.)

\noindent \textbf{Degree Discount} ~\cite{chen2009efficient}. The {\sf Degree Discount} (DDiscount) algorithm selects seed nodes based on their degree. Initially, the node with the highest degree is chosen. After selecting a seed, the degrees of its neighbors are adjusted to account for the influence of the already chosen seed. In each subsequent iteration, the node with the highest adjusted degree is selected as the next seed. A variation of DDiscount, the {\sf Single Discount} (SDiscount) algorithm, ensures that the connectivity of direct neighbors decreases by one for each chosen seed. This adjustment ensures that a node's influence isn't counted twice, providing a nuanced approach to seed selection in influence maximization.

\noindent \textbf{IMM} ~\cite{tang2015influence}. {\sf IMM} distinguishes itself by providing a guaranteed approximation ratio while maintaining efficiency, particularly suited for large-scale networks. The algorithm follows a two-phase approach. Firstly, it utilizes the {\sf Reverse Influence Sampling} (RIS) method to generate samples, assessing the reachability of nodes. Subsequently, in a greedy fashion, it judiciously selects seed nodes using these samples to maximize influence.

\begin{table*}[h]
    \centering
    \caption{Summary of datasets (K=$10^3$ M = $10^6$, B = $10^9$). Datasets marked with $\ast$ are only used in LND edge weight model.}
    \vspace{-3mm}
    \includegraphics[width=\linewidth]{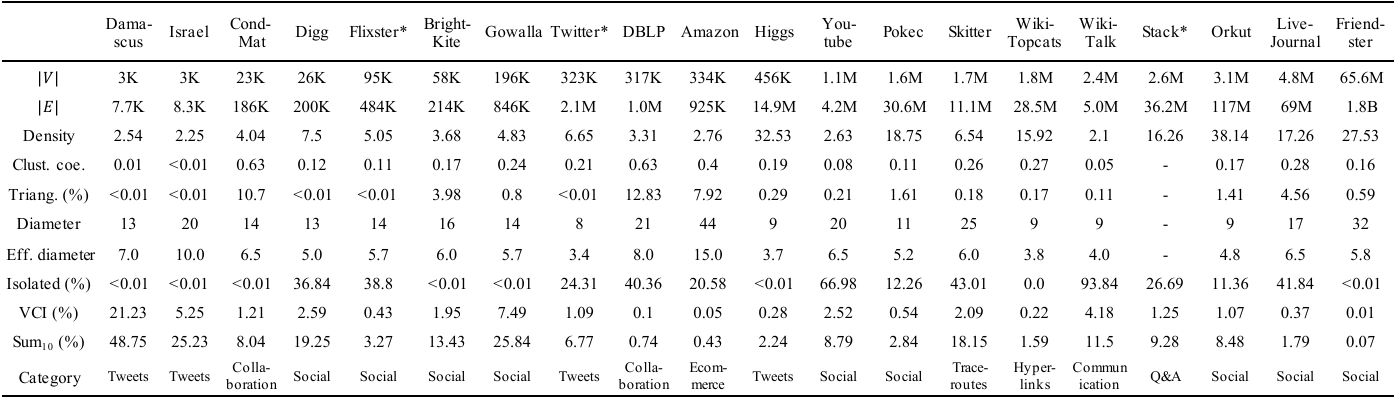}
    \vspace{-6mm}
    \label{dataset}
\end{table*}


\vspace{-2mm}
\subsection{Concerns on Training in Deep-RL Methods}\label{sec:concern}

In {\em all} the aforementioned Deep-RL methods~\cite{khalil2017learning,chen2021contingency,kamarthi2020influence,manchanda2020gcomb,ireland2022lense}, it's crucial to note that the training time for Deep-RL models is typically not considered in the computation cost. Treating the training time as a form of \textbf{pre-processing} might present a concern, as it could potentially affect the fairness of computational performance assessments.

In rigorous database research, it is common to amortize pre-processing time into the computation time of subsequent queries. While one could argue that not counting the training time is justifiable if the training dataset size is fixed and independent of subsequent MCP or IM queries, it's important to recognize that, theoretically, such pre-processing/training step takes \textbf{constant} time. Despite this constant pre-processing time, the trained model is expected to provide performance enhancements across various queries. However, it's worth noting that, theoretically, such Deep-RL-based methods may face challenges overcoming the $1-1/e$ approximation barrier. The rationale lies in the fact that constant preprocessing time, when added to polynomial query time, still results in polynomial time. According to complexity theory, no polynomial time algorithm can achieve an approximation ratio better than $1-1/e$ unless $\mathbf{P} = \mathbf{NP}$~\cite{kempe2003maximizing}. This insight raises considerations about the inherent limitations of these methods in surpassing certain approximation thresholds.


One may argue that the approximation ratio is w.r.t. the worst-case performance on all possible query graphs, while a trained Deep-RL model should be used to answer MCP or IM queries {\em with input graphs following the same distribution as the training graphs}. However, what does ``\textbf{same distribution}'' mean for graphs as input for MCP or IM? Can we use some easy-to-compute statistics of graphs to decide whether a testing graph is suitable for the trained Deep-RL model before running the model?

Another issue of the training of these Deep-RL methods is that the size of training data is \textbf{independent} to the future MCP or IM queries. Even though magically we can guarantee that the training graphs and testing graphs follow the ``same distribution'', according to basic statistical learning theory~\cite{mohri2018foundations}, the size of the training set has a crucial impact on the generalizability of the trained model. Therefore, a better way is to vary the size of the training data based on the MCP/IM queries we want to answer in the future.

Motivated by our above concerns, we conduct a thorough benchmark study to comprehensively examine the effectiveness and efficiency of the recent Deep-RL methods for MCP and IM~\cite{khalil2017learning,chen2021contingency,kamarthi2020influence,manchanda2020gcomb,ireland2022lense}.

\vspace{-3mm}
\subsection{Concerns on Lacking Strong Baselines} \label{sec:lazy}
\vspace{-0.8mm}

Lazy Greedy, an enhanced version of the conventional greedy algorithm, stands out for its remarkable efficiency and efficacy in MCP. However, this straightforward algorithm was neglected in studies of Deep-RL methods for MCP~\cite{ireland2022lense, khalil2017learning}.

%% file: 4a-Exp_Basic.tex
\vspace{-4mm}
\section{Benchmarking}\label{sec:benchmark}
\vspace{-0.8mm}
All the experiments were run on a server with 16 Intel i7-11700KF 3.60GHz cores, 64G RAM, and 1 NVIDIA GeForce RTX 3090 24G GPU. Our source code and datasets can be found at~\cite{LYKXG24Code}.

\noindent \textbf{Datasets.}
We selected datasets featuring diverse interaction and social structures essential for evaluating MCP and IM algorithms. This choice ensures a comprehensive analysis across varied network complexities. We utilized a set of 20 well-established real-world benchmark datasets~\cite{snapnets,manchanda2020gcomb}, which are commonly employed in existing studies, and their topology statistics are comprehensively outlined in Tab.~\ref{dataset}. The considered statistics encompass (1) the number of nodes $|V|$, (2) the number of edges $|E|$, (3) the graph density, (4) the average clustering coefficient~\cite{luce1949method}, (5) the fraction of closed triangles~\cite{watts2004six}, (6) the 90-percentile effective diameter~\cite{west2001introduction}, (7) the proportion of isolated nodes (those without neighbors), (8) the vertex centralization index, denoting the ratio of the maximum degree to the number of nodes, and (9) the Sum$_{10}$, representing the total degree of the top-10 nodes in the graph. We examine the strength and direction of association between these statistics and the performances of DeepRL methods in \S~\ref{sec:distribution}.

Our evaluations for MCP were conducted on 17 of these datasets. For the IM experiments, we conducted extensive evaluations across 10 datasets, employing edge weight models such as TV, CONST, and WC. In the case of the LND model, influence probabilities were generated using the credit distribution model \cite{goyal2011data} on Flixster and Twitter datasets. Additionally, the Stack dataset, sourced from \cite{manchanda2020gcomb}, was included in our experiments. To address scalability and performance challenges, {\sf RL4IM} was tested on small synthetic graphs using the power-law model \cite{onnela2007structure}. Furthermore, {\sf Geometric-QN} underwent evaluation on the small datasets mentioned in \cite{kamarthi2020influence}.


\begin{figure*}[h]
  \centering
  \vspace{-1.5mm}
  \includegraphics[width=0.95\linewidth]{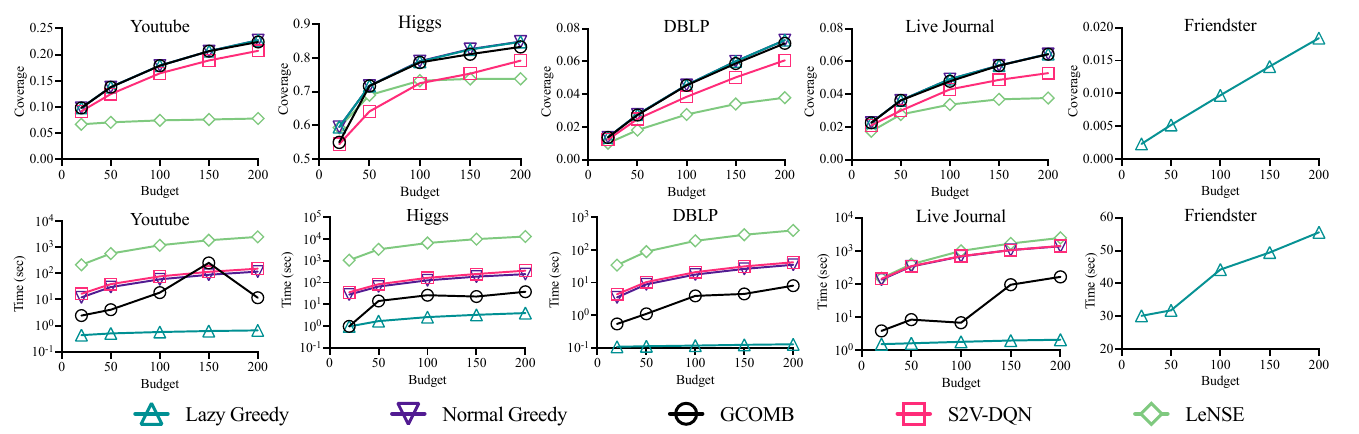}
  \vspace{-3mm}
  \caption{MCP: Coverage and Runtime curve}
  \label{fig:MCP}
  \vspace{-3mm}
\end{figure*}
 
\noindent \textbf{Implementation and hyperparameters setting.}
We implemented the {\sf Degree Discount} heuristic \cite{chen2009efficient} and the {\sf Lazy Greedy} algorithm \cite{leskovec2007cost}. For other methods, we utilized the code provided by the respective authors. All parameters were set consistently with the recommendations in prior studies \cite{tang2015influence,khalil2017learning,chen2021contingency,kamarthi2020influence,manchanda2020gcomb,ireland2022lense}. Specifically, we set $\epsilon$ to 0.5 for {\sf IMM} and 0.1 for {\sf OPIM}\footnote{In GCOMB~\cite{manchanda2020gcomb}, $\epsilon$ was set to 0.05 for {\sf OPIM}. However, setting $\epsilon=0.1$ still can guarantee a meaningful approximation ratio and we will show that {\sf OPIM} can still return high-quality solutions in such a case.}, respectively.

We adhered to the methodologies outlined in \cite{khalil2017learning,chen2021contingency,kamarthi2020influence,manchanda2020gcomb,ireland2022lense} to conduct the training of Deep-RL models for both MCP and IM. In the case of MCP, {\sf S2V-DQN}, {\sf GCOMB}, and {\sf LeNSE} were trained using the BrightKite dataset and subsequently tested on other datasets. For IM, {\sf GCOMB} and {\sf LeNSE} were trained on a subgraph randomly sampled from Youtube, with 15\% of edges selected at random. The model was then tested on various datasets under all edge weight models, excluding LND due to the absence of action logs. Beyond training models on real datasets, we extended our approach by training {\sf RL4IM} on synthetic graphs due to its scalability issue, inspired by Chen \textit{et al.} \cite{chen2021contingency}. The best result obtained from different training settings was considered for evaluation. 


\subsection{Training Time for Deep-RL Methods}

For fairness, we imposed a 24-hour training cap, selecting the optimal checkpoint based on the validation dataset. The training time required to reach this checkpoint (utilizing the WC model for IM) and the number of queries ($k$=200) executable by traditional SOTA methods within this timeframe across four datasets are detailed in Tab.~\ref{table:training time}. The results highlight that, while Deep-RL methods undergo extended training periods, traditional SOTA methods can execute queries numbering in the tens of thousands during the same interval, even on graphs exceeding 100 million nodes.

\begin{table}[]
\footnotesize
\caption{Training time of each method and number of queries answered by traditional methods ({\sf Lazy Greedy} for MCP and {\sf IMM} for IM) on four datasets in the same training time.}
\vspace{-2mm}
\begin{tabular}{cccccc}
\hline
Method       & Training     & \multicolumn{4}{c}{\#Queries answered within training time} \\ \cline{3-6} 
             & Time (mins.) & Pokec     & Wiki Talk    & Live Journal    & Orkut    \\ \hline
S2V-DQN      & 432.4        & 29522     & 16111        & 12506           & 5750     \\
GCOMB-MCP    & 22.3         & 1523      & 831          & 645             & 297      \\
LeNSE-MCP    & 105.5        & 7203      & 3931         & 3051            & 1403     \\
GCOMB-IM     & 312.8        & 9669      & 44056        & 58467           & 53167    \\
LeNSE-IM\tablefootnote{The results for LeNSE are reported after efficiency optimization (Please refer to Appendix in our full paper).}     & 96.6         & 2986      & 13606        & 18056           & 16419    \\
RL4IM        & 76.3         & 2359      & 10746        & 14262           & 12969    \\
Geometric-QN & 17.8         & 550       & 2507         & 3327            & 3025     \\ \hline
\end{tabular}
\vspace{-3mm}
\label{table:training time}
\end{table}

\subsection{Performance on MCP}

In this section, we conduct a benchmark of {\sf Normal Greedy}, {\sf Lazy Greedy}, {\sf S2V-DQN}, {\sf GCOMB}, and {\sf LeNSE} for MCP. The selective results are depicted in Fig.~\ref{fig:MCP}. Additional results for the remaining 12 datasets can be found in Appendix of our extended version~\cite{LYKXG24Code}.

\noindent \textbf{Effectiveness.} 
Both {\sf Normal Greedy} and {\sf Lazy Greedy} yeild an ($1-\frac{1}{e}$)-approximation solution. As evidenced by the results in Fig.~\ref{fig:MCP}. The performance of these two methods is comparable. {\sf GCOMB} outperforms {\sf S2V-DQN}, sometimes approaching or reaching the level of the greedy algorithms, consistent with the findings in \cite{manchanda2020gcomb}. However, other Deep-RL methods exhibit significantly poorer performance compared to Greedy. 
While {\sf GCOMB} generally performs closely to {\sf Lazy Greedy}, there are instances (e.g., on Digg, Skitter, and Higgs) where {\sf Lazy Greedy} still considerably outshines {\sf GCOMB}. However, we do not observe any marked distinction in the graph characteristics among these datasets~\cite{snapnets}, referring to the discussion about the ``same graph distribution'' claim (\S~\ref{sec:distribution}).


\begin{figure*}[h]
  \centering
  \vspace{-1mm}
  \includegraphics[width=0.95\linewidth]{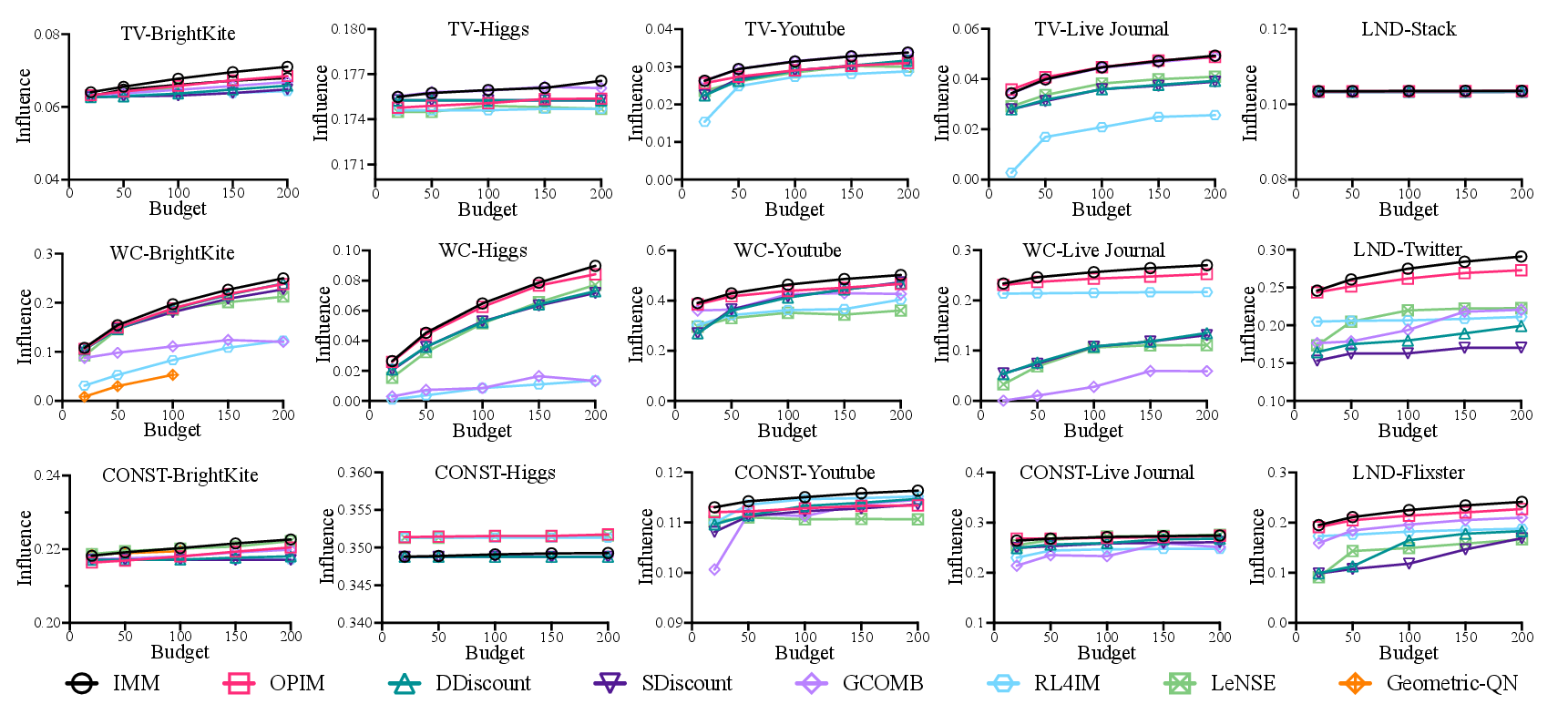}
  \vspace{-2mm}
  \caption{IM: influence curve under different weight models}
  \label{fig:im_effectiveness}
  \vspace{-2mm}
\end{figure*}

\begin{table}[]
\footnotesize
\caption{Memory usage (Gbyte). The upper part of the table records the peak memory usage of the algorithms in the MCP experiment, whereas the lower part records the usage in the IM experiment for the datasets BrightKite (BK), Youtube (YT), and Pokec (PK) under the WC, TV, and CONST (CO) models}
\vspace{-2mm}
\begin{tabular}{cccccc}
              \toprule
              & Gowalla & Youtube & Higgs & Pokec & Wiki Talk \\
              \midrule
S2V-DQN       & 0.58    & 2.12    & 6.47  & 11.96 & 3.69      \\
GCOMB         & 0.91    & 3.38    & 9.61  & 17.95 & 6.05      \\
LeNSE         & 0.78    & 3.00    & 8.53  & 15.1  & 5.45      \\
Lazy Greedy   & 0.18    & 0.71    & 1.61  & 3.23  & 1.28      \\
Normal Greedy & 0.01    & 0.03    & 0.01  & 0.04  & 0.06     \\
             \midrule
              & BK-WC   & BK-TV   & YT-CO & PK-WC & PK-CO     \\ 
              \midrule
IMM           & 0.02    & 0.38    & 0.41  & 0.84  & 27.18     \\
OPIM          & 0.02    & 0.23    & 0.18  & 0.43  & 19.00     \\
DDiscount     & 0.01    & 0.02    & 0.14  & 0.64  & 0.64      \\
LeNSE         & 0.34    & 0.34    & 3.30  & 20.3  & 19.99     \\
GCOMB         & 2.15    & 1.43    & 3.80  & 13.77 & 13.47     \\
RL4IM         & 0.05    & 0.05    & 0.69  & 3.93  & 3.89      \\
Geometric-QN & 0.28    & 0.3559  & /     & /     & /         \\
\bottomrule
\end{tabular}

 \label{table:memory}
 \vspace{-3mm}
\end{table}

\noindent \textbf{Efficiency.}
Fig.~\ref{fig:MCP} illustrates the efficiency of {\sf GCOMB}, which is 1 to 2 orders of magnitude faster than {\sf S2V-DQN} and over 2 orders faster than {\sf LeNSE} in most scenarios. Particularly with a small budget, it surpasses the runtime of {\sf Normal Greedy} by up to two orders, consistent with the findings in \cite{manchanda2020gcomb}. 
The runtime of {\sf GCOMB} exhibits fluctuations rather than a steady increase, primarily attributed to the varying number of good nodes predicted by its node pruner
\footnote{For a more comprehensive experimental analysis, refer to the detailed results in the Appendix of our full paper}. In contrast, {\sf LeNSE} takes over 10$\times$ longer than {\sf Normal Greedy}.

{\sf Lazy Greedy} runs more than one order of magnitude faster than {\sf GCOMB} when the budget is small, and this gap widens to up to two orders of magnitude as the budget increases. Additionally, we conducted tests on a billion-sized graph, Friendster. {\sf Lazy Greedy} successfully solves the problem within minutes. In contrast, {\sf GCOMB} crashes in our experimental environment, so we compare it with the result reported in \cite{manchanda2020gcomb}, which is two orders of magnitude slower than {\sf Lazy Greedy}. Note that \textbf{the runtime of Deep-RL methods only counts inference time}, excluding the extra time of the preprocessing and training phases. Despite this, even when Deep-RL methods are given some unfair advantages in the comparison, {\sf Lazy Greedy} consistently outperforms all Deep-RL methods.


\noindent \textbf{Memory Usage.} Tab.~\ref{table:memory} provides insights into the memory consumption of each method across representative datasets. In the inference phase, Deep-RL methods exhibit a memory footprint at least 3$\times$ larger than {\sf Lazy Greedy} and 78$\times$ larger than {\sf Normal Greedy}. Importantly, Deep-RL algorithms typically impose even higher memory demands during the training phase.



\noindent \textbf{Summary.} Combining all the above results, we find that in our experiments for MCP, {\sf Lazy Greedy} dominates all Deep-RL methods on effectiveness, efficiency, and memory usage.


\vspace{-2mm}
\subsection{Performance on IM}
\label{sec:performance_im}

\begin{figure*}[h]
  \centering
  \vspace{-1mm}
  \includegraphics[width=.95\linewidth]{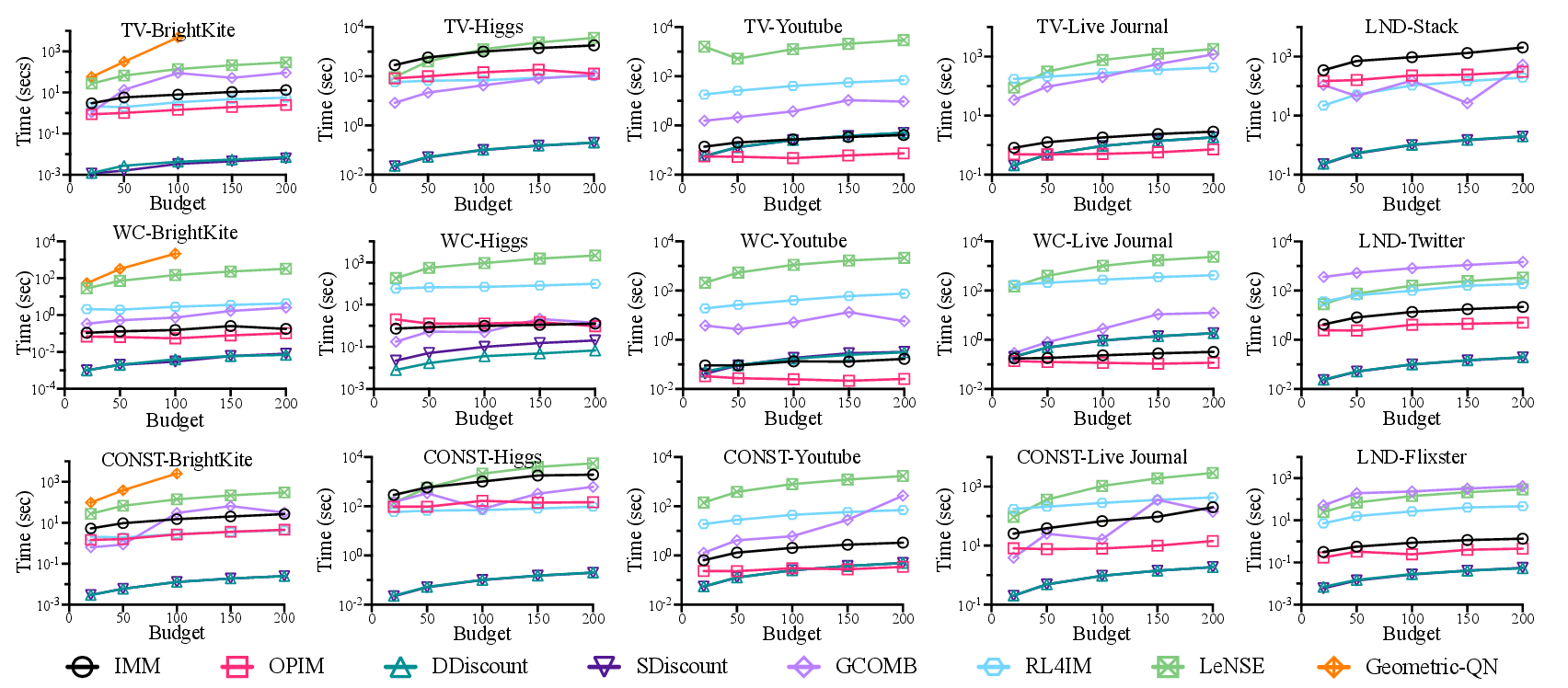}
  \vspace{-3mm}
  \caption{IM: runtime curves under different weight models.}
  \label{fig:im_efficiency}
  \vspace{-2mm}
\end{figure*}

\begin{figure}\centering
    \begin{subfigure}[b]{0.44\linewidth}
        \includegraphics[width=\linewidth,height=0.78\linewidth]{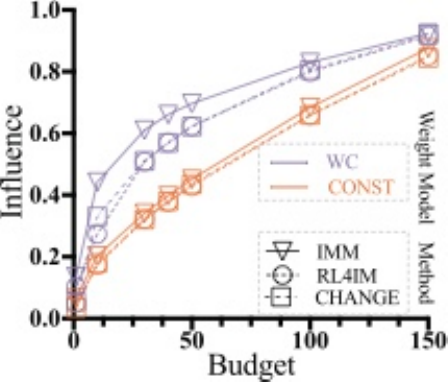}
        \vspace{-5mm} 
        \caption{RL4IM}
        \label{fig:rl4im}
    \end{subfigure}
    \begin{subfigure}[b]{0.54\linewidth}
        \includegraphics[width=\linewidth,height=0.68\linewidth]{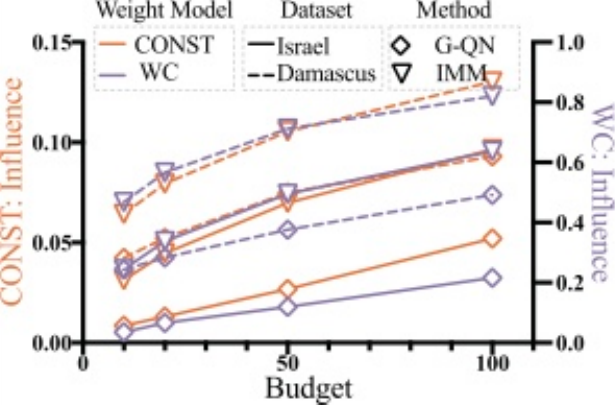}
        \vspace{-5mm}
        \caption{Geometric-QN}
        \label{fig:gqn}
    \end{subfigure}
    \vspace{-3mm} 
    \caption{(a) Comparison between {\sf RL4IM}, {\sf CHANGE}, and {\sf IMM} over synthetic graphs. The coverage is averaged of over 10 repeated experiments. (b) Comparison between {\sf Geometric-QN}(G-QN) and {\sf IMM} over small-scale datasets. The coverage is an average of the results of 20 repeated experiments.}
    \vspace{-4mm}
    \label{fig:reproduce_rl4im_syn}
\end{figure}

In this section, we assess the performance of Deep-RL methods, namely {\sf GCOMB}, {\sf RL4IM}, and {\sf Geometric-QN}, while also benchmarking traditional algorithms like {\sf IMM}, {\sf OPIM}, {\sf Degree Discount}, and {\sf Single Discount} in our IM experiments. 
All algorithms underwent testing across four edge weight models: CONST, TV, WC, and LND. It is noteworthy that none of the Deep-RL studies ~\cite{khalil2017learning,chen2021contingency,kamarthi2020influence,manchanda2020gcomb,ireland2022lense} have explored their methods under the WC model. This model, arguably the most prevalent in the IC model for IM literature, is included in our evaluation for comprehensive insights.




\noindent \textbf{Effectiveness.} As shown in Fig.~\ref{fig:im_effectiveness}, in line with the findings presented in \cite{manchanda2020gcomb}, {\sf GCOMB} performs comparably to {\sf IMM} on Youtube under TV and on Stack under LND. However, it slightly lags behind {\sf IMM} on Youtube under CONST. 
{\sf RL4IM} exhibits greater stability than {\sf GCOMB} and delivers a more effective solution, particularly under the CONST model. 
Despite being the most effective among the learning methods in various cases, {\sf LeNSE} still falls short when compared to classical algorithms. 
Notably, these learning methods {\em display limited effectiveness across different datasets}, suggesting {\bf poor generalizability}. 
It is worth highlighting that instances where Deep-RL methods match the effectiveness of {\sf IMM} are characterized by situations where {\em the influence spread does not increase with an expanding budget}. In such {\bf atypical cases}, the influence spread is {\em primarily governed by a few nodes}, making marginal increments subtle and challenging to observe. In such instances, {\sf IMM} may exhibit inefficiency due to the subtle differences in the marginal gain of nodes, necessitating the generation of numerous RR sets to distinguish potential candidate nodes. 
Furthermore, under the WC or LND model, Deep-RL methods consistently underperform {\sf IMM}.

We observe that {\sf IMM} remains the most effective algorithm, with {\sf OPIM} exhibiting similar effectiveness. Especially under WC model and LND model, the performance gap between theoretically sound algorithms ({\sf IMM} and {\sf OPIM}) and Deep-RL methods is especially prominent. 
Surprisingly, even discount algorithms, despite being heuristics, outperform Deep-RL methods across most cases. 
{\em These results cast doubt on the effectiveness of Deep-RL methods compared to traditional IM algorithms.}



\noindent \textbf{Additional Evaluation over Synthetic Datasets.} Given the observed poor performance of {\sf RL4IM} and {\sf Geometric-QN} on large-scale real-world datasets, coupled with the failure of {\sf Geometric-QN} on nearly all such datasets due to its intensive memory requirements, we conducted additional experiments to evaluate their effectiveness on smaller datasets as suggested in the respective papers. 
{\sf RL4IM} was trained following the instructions in \cite{chen2021contingency}, and all methods repeated the query over ten times, calculating the average result. In line with the findings in \cite{chen2021contingency}, Fig.~\ref{fig:reproduce_rl4im_syn} illustrates that {\sf RL4IM} outperforms {\sf CHANGE} on synthetic graphs with 200 nodes and a small budget. However, extending the experiment to larger graphs with 2000 and 20,000 nodes under CONST and WC models, {\sf RL4IM} consistently surpasses {\sf CHANGE} but falls short of {\sf IMM}. This indicates that {\em while {\sf RL4IM} performs well in small synthetic graphs, it remains inferior to {\sf IMM} in this context}.
Regarding {\sf Geometric-QN}, \cite{kamarthi2020influence} reported that it obtains 37.8\% and 61.1\% of the influence scores of the greedy algorithm for the datasets Israel and Damascus, respectively. As {\sf IMM} provides the same approximation ratio as the greedy algorithm, we directly compared {\sf Geometric-QN} with {\sf IMM} in our experiments. Due to the high variance of {\sf Geometric-QN}, we repeated the query over 20 times and then calculated the average as the final result. Our experiment showed that {\sf Geometric-QN} achieves 27.5\% and 66.1\% of the coverage of {\sf IMM} in Israel and Damascus, respectively.  Though unlike what \cite{kamarthi2020influence} reported, {\sf Geometric-QN}  unmistakably lags behind {\sf IMM}. 

\begin{table*}[]
\footnotesize
\caption{Correlation of Graph Metrics with Coverage Gap using Spearman's Coefficient: Highlighting Values $\geq$0.8. The upper section represents unweighted topological metrics, the middle section denotes weighted metrics, and the lower section outlines the complex metrics.}
\vspace{-2mm}
\begin{tabular}{@{}ccccccccccccc@{}}
\toprule
                             & & MCP         &        &         & CONST  &                &        & TV              &                 &                & WC              &               \\

                             & LeNSE  & GCOMB           & S2V-DQN & LeNSE  & GCOMB          & RL4IM  & LeNSE           & GCOMB           & RL4IM          & LeNSE           & GCOMB  & RL4IM  \\
\midrule
$|V|$                        & -0.286 & \textbf{0.943} & 0.771   & -0.710 & 0.143  & 0.257  & 0.721          & 0.714          & 0.371  & 0.7            & -0.143         & 0.486  \\
$|E|$                        & -0.238 & 0.543          & 0.314   & -0.912 & 0.371  & 0.543  & \textbf{0.901} & 0.6            & 0.657  & \textbf{0.870} & -0.429         & 0.257  \\
Density                      & 0.024  & -0.6           & -0.6    & 0.321  & 0.6    & 0.771  & 0.323          & 0.143          & 0.429  & 0.3            & 0.029          & -0.2   \\
Clust. coe. & 0.5    & -0.6           & -0.6    & -0.728 & -0.486 & -0.429 & -0.712         & -0.829         & -0.543 & -0.712         & -0.314         & -0.6   \\
Triang. (\%) & 0.381  & -0.714         & -0.657  & -0.772 & -0.543 & -0.543 & -0.689         & -0.886         & -0.657 & -0.692         & -0.086         & -0.543 \\
Diameter                     & 0.122  & -0.174         & -0.174  & -0.872 & -0.841 & -0.696 & -0.872         & -0.638         & -0.986 & -0.873         & 0.203          & -0.232 \\
Eff. diameter      & 0.072  & -0.086         & -0.086  & -0.921 & -0.886 & -0.771 & -0.901         & -0.6           & -1.0   & -0.901         & 0.257          & -0.143 \\
Isolated (\%)            & -0.18  & 0.783          & 0.522   & -0.112 & -0.464 & -0.261 & -0.120         & 0.464          & -0.145 & -0.1           & 0.319          & 0.754  \\
VCI (\%)                          & -0.524 & 0.486          & 0.371   & 0.652  & 0.429  & 0.486  & 0.6            & \textbf{0.886} & 0.6    & 0.6            & 0.429          & 0.771  \\
Sum$_{10}$ (\%)                        & -0.476 & -0.029         & 0.029   & 0.612  & 0.6    & 0.486  & 0.6            & 0.6            & 0.6    & 0.6            & 0.486          & 0.486  \\

\hdashline

weighted degree         & -      & -              & -       & 0.486  & 0.429  & 0.371  & 0.2            & -0.371         & 0.257  & 0.371          & -0.314         & -0.6   \\
edge weight             & -      & -              & -       & 0.371  & 0.257  & 0.371  & 0.143          & 0.543          & 0.257  & 0.486          & \textbf{0.829} & 0.657  \\

\hdashline

Community Structure        & -       & -               & - & \textbf{0.942} & \textbf{0.812} & \textbf{0.952} & \textbf{0.912} & \textbf{0.907} & \textbf{1.0} & 0.643 & -0.351 & 0.398 \\
WL kernel                   & -      & -              & -  & 0.621 & \textbf{0.882} & 0.636 & 0.515 & 0.135 & 0.321 & \textbf{0.922} & 0.0 & -0.523 \\
PageRank                     & -      & -              & - & -0.653 & -0.716 & -0.636 & -0.475 & 0.132 & -0.334 & -0.653 & 0.366 & 0.73 \\

\bottomrule
\end{tabular}
\label{table:distribution spearman}
\vspace{-2mm}
\end{table*}

\noindent \textbf{Efficiency.} It is important to note that we provide unfair advantages to Deep-RL methods by excluding their pre-processing or training time. Despite this, as shown in Fig.~\ref{fig:im_efficiency}, traditional IM algorithms are still 10$\times$ to 10,000$\times$ faster than Deep-RL methods in most cases. However, in scenarios where the influence spread hardly increases with the budget, such as in Pokec, Wiki Talk, and Wiki Topcats under the CONST model, there are numerous solution sets with very similar influence spread, i.e., the atypical cases discussed in the effectiveness assessments. Distinguishing these highly similar solution sets requires generating many RR sets, making theoretically sound algorithms like {\sf IMM} and {\sf OPIM} slow in such situations.


With the help of node pruning techniques, {\sf GCOMB} can sometimes achieve speeds that are orders of magnitude faster than {\sf IMM}. However, the runtime of {\sf GCOMB} is often non-monotonic concerning the budget, indicating that the node pruner cannot guarantee a smaller search space for a small budget compared to a large budget (e.g., TV-BrightKite, CONST-Higgs, CONST-Live Journal). The instability of the node pruner leads to {\sf GCOMB} being orders of magnitude slower than {\sf IMM} in certain cases.(eg. WC-Youtube, TV-Live-Journal, and more cases displayed in our full version).

In contrast to {\sf GCOMB}, which uses a simple linear interpolation method to estimate pruning thresholds, {\sf LeNSE} iteratively constructs subgraphs in a Markov decision process manner to achieve pruning effects. This makes {\sf LeNSE} significantly slower than other methods and incapable of finding solutions for large datasets like Orkut and Stack within a limited time. {\sf Geometric-QN} employs a computationally expensive real-time graph exploration policy during the inference process, rendering it even non-scalable to moderate-size Higgs dataset.




\noindent \textbf{Memory Usage.}
The lower section of Tab.~\ref{table:memory} presents the peak memory usage of algorithms in the IM experiments. Memory consumption by Deep-RL methods varies significantly. {\sf Geometric-QN} is memory-efficient, but it takes considerably more time to find a solution, leading to poor scalability. As discussed in the efficiency analysis, {\sf IMM} and {\sf OPIM} need to generate a large number of RR sets in atypical scenarios, resulting in substantial memory usage. Conversely, in other situations, {\sf IMM} and {\sf OPIM} tend to consume less memory than Deep-RL methods.

\noindent \textbf{Summary.} Based on the above evaluations, we find that except for the weird cases when the influence spread is insensitive to the increasing budget, {\em traditional IM algorithms outperform Deep-RL methods in effectiveness, efficiency, and memory usage.}

%% file: 4b-Exp_limit.tex
\section{common issues of Deep-RL methods}\label{sec:deeper}
The results presented in \S~\ref{sec:benchmark} contradict the expectations of all the Deep-RL methods~\cite{khalil2017learning,chen2021contingency,kamarthi2020influence,manchanda2020gcomb,ireland2022lense}, which aim to efficiently learn better approximations of the coverage function or the influence function, leading to more effective and efficient solutions for MCP or IM. 
Given the notable performance disparity between Deep-RL methods and {\sf Lazy Greedy} in MCP, this section shifts its focus to examine the application of Deep-RL in IM. Additional experiments are conducted to unveil prevalent issues within the Deep-RL methods~\cite{khalil2017learning,chen2021contingency,kamarthi2020influence,manchanda2020gcomb,ireland2022lense} that hinder their practical effectiveness.

\begin{table}[]
\footnotesize
\caption{Percentage change of the performance}
\vspace{-2mm}
\begin{tabular}{@{}ccccccc@{}}
\toprule
           & \multicolumn{3}{c}{TV} & \multicolumn{3}{c}{WC} \\ \midrule
           & GCOMB   & RL4IM   & LeNSE    & GCOMB   & RL4IM   & LeNSE    \\
BrightKite & 1.29\%  & -0.95\% & -0.19\%  & 30.18\% & 26.52\% & -11.73\% \\
Amazon     & 21.74\% & 27.23\% & -7.69\%  & 7.64\%  & 65.52\% & 36.45\%  \\
DBLP       & 86.81\% & 43.58\% & -18.18\% & -5.49\% & 87.61\% & 43.77\%  \\
Wiki Talk  & 2.46\%  & -2.26\% & 7.59\%   & 6.39\%  & 83.10\% & 10.20\%  \\
Youtube    & 59.22\% & 13.26\% & 3.18\%   & 0.61\%  & 2.80\%  & 3.33\%   \\ \bottomrule
\end{tabular}
\label{table:distribution gap}
\vspace{-2mm}
\end{table}

\subsection{Study of Graph Distribution}
\label{sec:distribution}

In \S~\ref{sec:concern}, we delve into the limitations posed by the worst-case performance of Deep-RL methods, highlighting the {\em theoretical boundary} of $1-\frac{1}{e}$. Moreover, we emphasize that the practical performance of these methods in scenarios beyond the worst-case is contingent on their {\em generalizability}.
Deep-RL studies~\cite{khalil2017learning, manchanda2020gcomb, chen2021contingency, kamarthi2020influence, ireland2022lense} assert that Deep-RL methods can learn heuristics to solve combinatorial problems on graphs and generalize effectively to graphs resembling the training distribution. This assertion prompts an evaluation of the similarity between the distributions of the training and test graphs.
However, these studies {\em lack a rigorous definition of ``graph distribution''}. In practical terms, using trained Deep-RL models for MCP and IM requires an efficient method to determine whether a new input graph conforms to the ``same distribution'' as the training graphs. We explore the feasibility of leveraging {\em easy-to-compute statistics of graphs} to ascertain if two graphs follow the ``same distribution''.

\noindent \textbf{Edge Weights Matter in IM.} 
We begin by highlighting the significance of edge weights in IM. Our investigation aims to assess the generalization capability of trained Deep-RL models when applied to {\em the same graph under edge-weight models different from the model used during training}.
To illustrate, let $G_{M}$ be a graph using the edge weight model $M$, and $\mathcal{F}_M$ be the Deep-RL method $\mathcal{F}$ trained on $G_M$. Adhering to the experimental setup from baseline papers, we select CONST as the training model. 
Subsequently, we evaluate the performance of $\mathcal{F}_{CO}$ against $\mathcal{F}_M$ across five graphs, utilizing the edge weight model $M$ with a budget of $k=50$. Tab.~\ref{table:distribution gap} lists the percentage change of the performance $p$, where $p=\frac{\mathcal{F}_{M}(G_M) - \mathcal{F}_{CO}(G_M)}{\mathcal{F}_{M}(G_M)}$. Here, $\mathcal{F}_{CO}(G_M)$ means $\mathcal{F}$ is trained under CONST while tested on $G_M$, with $M \in \{WC, TV\}$. 
A larger absolute value indicates greater sensitivity of the method $\mathcal{F}$ to the edge weight model. The results suggest that these Deep-RL methods {\em struggle to generalize effectively across different edge-weight models}.

\begin{figure*}\centering
    \begin{subfigure}[b]{0.24\linewidth}
        \includegraphics[width=\linewidth,height=0.7\linewidth]{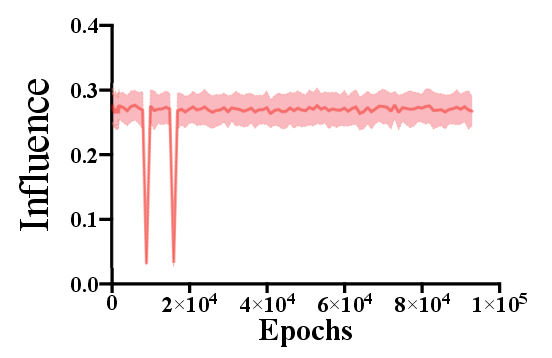}
        \vspace{-4mm}
        \caption{GCOMB}	
        \label{fig:gcomb_epoch}
    \end{subfigure}
    \begin{subfigure}[b]{0.24\linewidth}
        \includegraphics[width=\linewidth,height=0.7\linewidth]{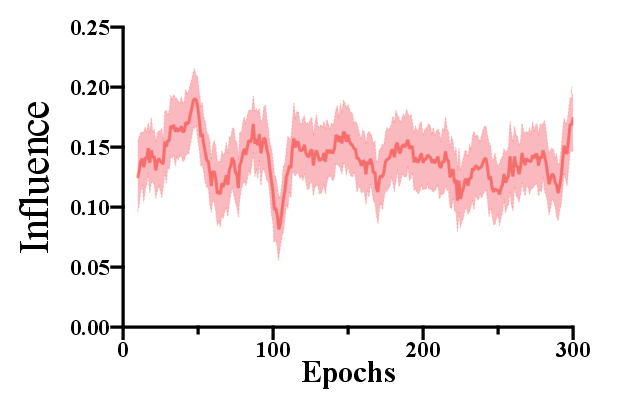}
        \vspace{-4mm}
        \caption{LeNSE}	
        \label{fig:lense_epoch}
    \end{subfigure}
    \begin{subfigure}[b]{0.24\linewidth}
        \includegraphics[width=\linewidth,height=0.7\linewidth]{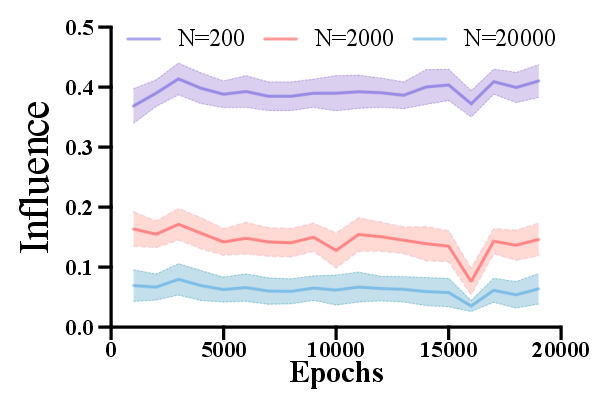}
        \vspace{-4mm}
        \caption{RL4IM}
        \label{fig:rl4im_epoch}
    \end{subfigure}
    \begin{subfigure}[b]{0.24\linewidth}
        \includegraphics[width=\linewidth,height=0.7\linewidth]{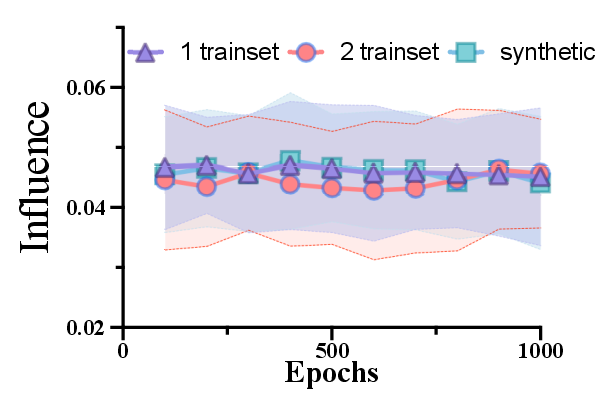}
        \vspace{-4mm}
        \caption{Geometric-QN}
        \label{fig:gqn_epoch}
    \end{subfigure}
    \vspace{-3mm}
    \caption{Performance Curves of Models with Varying Training Durations under the WC Model. (a-b): Training and validation datasets are 15\% subgraphs from YouTube, with the test dataset from the remaining 70\% edges, as in \cite{manchanda2020gcomb}. (c): Training on 200 synthetic graphs (200 nodes each); testing on synthetic graphs of varying number ($N$) of nodes. (d): Following \cite{kamarthi2020influence}, models are trained on different datasets, including the Copen dataset (1 trainset), both Copen and Occupy datasets (2 trainset), and synthetic graphs (synthetic)}
    \label{fig:time}
    \vspace{-3mm}
\end{figure*}

\begin{table}[]
\small
\caption{Ratio of runtime: Advanced graph similarity calculation approaches to OPIM query resolution with $k=200$}
\vspace{-2mm}
\begin{tabular}{@{}ccccccc@{}}
\toprule
          & \multicolumn{3}{c}{DBLP} & \multicolumn{3}{c}{Wiki Talk} \\ 
          & CONST   & TV    & WC     & CONST    & TV      & WC       \\ \midrule
Community & 1625.5  & 371.7 & 1730.9 & 45.4     & 56.6    & 1851.5   \\
WL Kernel & 258.0   & 73.3  & 382.1  & 4.3      & 5.5     & 243.2    \\
PageRank  & 79.8    & 22.9  & 120.8  & 2.3      & 2.7     & 120.5    \\ \bottomrule
\end{tabular}
\vspace{-2mm}

\label{table:distribution runtime}
\end{table}


\noindent \textbf{Common topology statistics do not help.} Next, we investigate whether commonly adopted topology statistics of graphs can help discern whether a testing graph follows the ``same distribution'' as the training graphs.
In this context, we employ the same edge-weight model for both training and testing data. In addition to unweighted topological statistics (1)-(9) mentioned in the dataset introduction, for IM, we also incorporate topological metrics associated with edge weights, such as (10) the average weighted degree and (11) the average edge weight.

We calculate the Spearman correlation coefficient\footnote{The Spearman correlation coefficient is a widely-adopted metric to measure the strength and direction of association between two ranked variables~\cite{spearman1961proof}.} between the coverage gap $\mathcal{\delta}^{n \times 1}$ and $m$ topology statistics of graphs $\mathcal{X}^{n \times m}$ across $n$ datasets. Here, $\delta_i = \frac{\mathcal{F(\cdot)}_i - OPT_i}{OPT_i}$, where $OPT_i$ is approximated by the coverage obtained over dataset $i$ by greedy in MCP and IMM in IM, and $\mathcal{F}(\cdot)$ is the coverage or influence obtained by a Deep-RL method $\mathcal{F}$. Tab.~\ref{table:distribution spearman} shows the results. Unfortunately, {\em strong positive correlations ($\geq 0.8$) are relatively rare}. What's worse, even when a statistic has a strong positive correlation with a particular method, the strong correlation will no longer exist once we change the edge-weight model (e.g., the VCI(\%) for {\sf GCOMB} TV and WC).

\noindent \textbf{Derivation of Complex Topological Statistics is Computationally Intensive.} In addition to straightforward and easy-to-compute topological statistics of graphs, we delve into intricate methods like Weisfeiler-Lehman (WL) Graph Kernel \cite{shervashidze2011weisfeiler}, PageRank, and Louvain community detection \cite{blondel2008fast} to ascertain similarity for weighted graphs in IM. Insights from Tab.~\ref{table:distribution spearman} illustrate that the {\em WL graph kernel and PageRank are markedly inefficient in pinpointing graphs with analogous distributions}. On the other hand, {\em analogous community structures effectively discern similar distributions in the TV and CONST models}. However, mirroring the outcomes observed in learning approaches, community structures under the WC model fall short of capturing the similarities between graphs. This indicates the potential significance of community structures as pivotal indicators while assessing similar distributions {\em prior to inference}. Nevertheless, the precision offered by these advanced metrics comes with the trade-off of being impractical for real-time or large-scale applications due to their {\em substantial computational overhead}. Tab.~\ref{table:distribution runtime} presenting results from both a small and a large dataset, accentuates that the computation of these metrics far exceeds the time frame required by the state-of-the-art {\sf OPIM} to resolve a query when $k=200$. 
It's noteworthy that, in practical scenarios, {\em IM queries do not necessarily demand high throughput}. Identifying seed sets for campaigns in a social network, for instance, typically involves a manageable number of candidates rather than thousands or hundreds of thousands.

\noindent {\bf Summary.} These evaluations imply that determining whether a testing graph aligns with the ``same distribution'' as the training graphs might be challenging in practical applications. Consequently, assessing the performance of a trained Deep-RL model on a testing graph may require actual execution on the specific graph, as predicting its effectiveness beforehand could be a non-trivial task.

\subsection{Impact of Training Time}

Deep-RL methods frequently encounter convergence challenges in practical scenarios~\cite{arulkumaran2017deep}. To investigate the convergence behavior of Deep-RL methods for MCP and IM~\cite{khalil2017learning, manchanda2020gcomb, chen2021contingency, kamarthi2020influence, ireland2022lense}, we conduct experiments by keeping the training data constant while varying the training time (epochs).  
It is essential to note that we extend the training duration significantly beyond that reported in the original papers and select the best checkpoint based on the validation set. Despite these efforts, the Deep-RL methods consistently exhibit inferior performance compared to {\sf IMM}. 
{\sf GCOMB} (Fig.~\ref{fig:gcomb_epoch}) initially has an unusual performance: drop but then tends to converge as the number of training epochs increases. 
{\sf LeNSE} (Fig.~\ref{fig:lense_epoch}) demonstrates effective learning initially but encounters intermittent performance drops. 
{\sf RL4IM} (Fig.~\ref{fig:rl4im_epoch}) exhibits a steady performance improvement until approximately 3,000 epochs, beyond which no further enhancement is observed, indicating potential overfitting and diminished generalizability with prolonged training.
Similarly, {\sf Geometric-QN} (Fig.~\ref{fig:gqn_epoch}) fails to learn efficiently as the training steps increase. 
In summary, extending the training time for Deep-RL methods may not consistently lead to improved performance, posing challenges in determining an optimal training duration and making the tuning process intricate in practical applications.

\subsection{Impact of Training Dataset Size}

\begin{figure*}[h]
    \begin{subfigure}[b]{\linewidth}
        \begin{subfigure}[b]{0.33\linewidth}
		  \includegraphics[width=0.9\linewidth,height=0.6\linewidth]{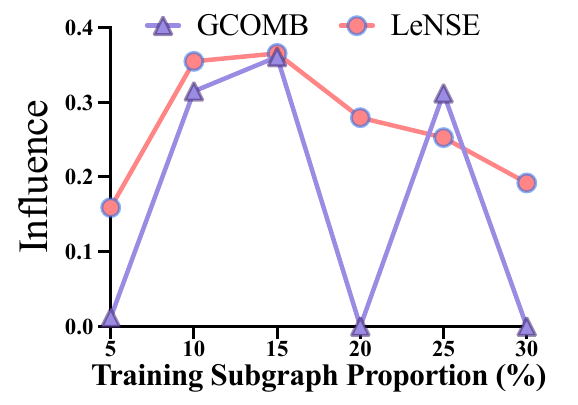}
            \vspace{-1mm}
		  \caption{GCOMB \& LeNSE}
		  \label{fig:GL_trainsize}
		\end{subfigure}
		\begin{subfigure}[b]{0.33\linewidth}
		  \includegraphics[width=0.9\linewidth,height=0.6\linewidth]{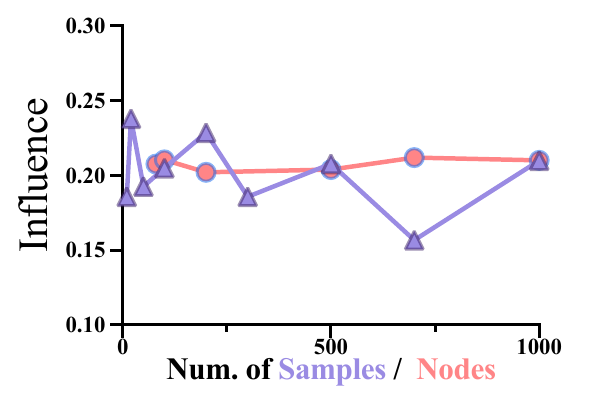}
            \vspace{-1mm}
		  \caption{RL4IM}
		  \label{fig:RL4IM_trainsize}
		\end{subfigure}
		\begin{subfigure}[b]{0.33\linewidth}
		  \includegraphics[width=0.9\linewidth,height=0.6\linewidth]{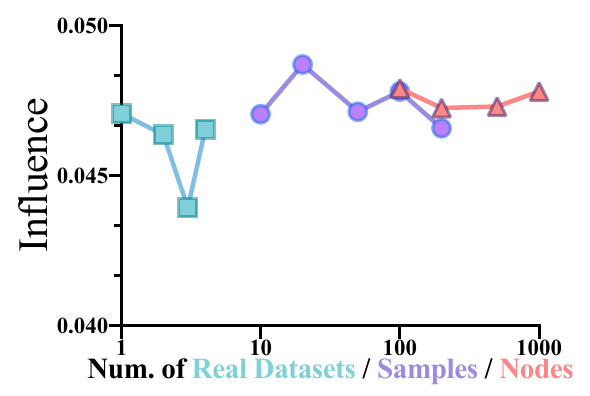}
            \vspace{-1mm}
		  \caption{Geometric-QN}
		  \label{fig:gqn_trainsize}
		\end{subfigure}
	\end{subfigure}
    \vspace{-5mm}
	\caption{Performance Curves of Models under the WC Model Based on Training Dataset Size. (a): x-axis represents the percentage of YouTube edges used for training. (b): x-axis denotes the number of \textcolor[RGB]{151,142,221}{samples} (200 nodes each) or \textcolor[RGB]{240,139,138}{nodes} (200 samples total). (c): x-axis indicates the number of synthetic \textcolor[RGB]{151,142,221}{samples}, \textcolor[RGB]{240,139,138}{nodes} in a synthetic graph, or the number of \textcolor[RGB]{145,206,214}{real datasets}.}
 \label{fig:size}
 \vspace{-1mm}
\end{figure*}

In this study, we address the impact of training dataset size on the performance of Deep RL methods, acknowledging the challenges associated with acquiring datasets with high-quality labels. The training approaches for Deep RL methods in our investigation fall into two categories based on the size of the training datasets. 
{\sf GCOMB} and {\sf LeNSE} are trained on fractional subsets of a dataset, while {\sf RL4IM} and {\sf Geometric-QN} are trained on multiple datasets. Consequently, the training dataset size can refer to either the number of nodes in a graph or the number of graphs (i.e., samples).
To systematically explore the influence of training dataset size, we adopt distinct strategies for each category. For {\sf GCOMB} and {\sf LeNSE}, we train these models using various subgraphs, each comprising no more than 30\% of the nodes in the Youtube dataset. Subsequently, we evaluate the models on the same graph, constructed with the remaining 70\% of edges unseen during the training phase.
Following the methodology outlined in \cite{kamarthi2020influence}, we train {\sf Geometric-QN} on different numbers of real datasets or synthetic graphs. In the case of {\sf RL4IM} and {\sf Geometric-QN} trained on synthetic graphs, we not only vary the number of samples while keeping the number of nodes fixed but also train the models with a fixed number of samples, each having different numbers of nodes. 

The experimental results reveal that {\em none of the methods consistently show improved performance with an increased training dataset size}. 
More specifically, from the effect of the number of nodes in the training subgraphs on the model's performance, Fig.~\ref{fig:GL_trainsize} shows that {\sf GCOMB} and {\sf LeNSE} achieves their optimal performances when trained on a subgraph of 15\% size, but their performance drastically drops as the training dataset size increases, indicating instability. 
In contrast, Fig.\ref{fig:RL4IM_trainsize} reveals that {\sf RL4IM} displays more stability, with a subtle trend suggesting that {\sf RL4IM} tends to perform better as the size of the training subgraphs increases when tested on graphs of similar size. 
However, there is no significant correlation between generalizability and training dataset size, as {\sf RL4IM} trained on smaller graphs (n=100) outperforms the model trained on larger graphs when tested on graphs that are 10 or 100 times the size of the training graphs. 
Examining the impact of the number of training subgraphs, size does not significantly affect {\sf RL4IM}'s performance or generalizability.
Furthermore, Fig.~\ref{fig:gqn_trainsize} indicates that increasing the number of training graphs does not improve the performance of {\sf Geometric-QN} either. 
These results underscore the critical insight that {\em simply enlarging the training dataset size may have detrimental effects on the model's performance}. Therefore, determining {\em an appropriate volume of training data} 
poses a significant challenge.


\section{Rating Scale for Each Solver}

Fig.~\ref{fig:overview_performance} in \S~\ref{sec:intro} illustrates the relationship between the average normalized coverage or influence (y-axis) and the average normalized runtime (x-axis) across 16 datasets in MCP and 8 datasets under three weight models in IM. A position closer to the top-left corner indicates a more desired performance -- faster and more effective. Higher robustness is suggested by a lower standard deviation value in effectiveness, while a higher value indicates the opposite.  
Among Deep-RL methods in MCP, {\sf GCOMB} offers coverage comparable to greedy algorithms and outpaces the standard greedy approach. However, it consistently falls short when compared to {\sf Lazy Greedy}. 
In IM, Deep-RL methods even lag behind simple heuristic methods like {\sf SDiscount} and {\sf DDiscount}. 
It's noteworthy that the presented runtimes exclude the preprocessing and training time of these Deep-RL methods. Despite this, they remain significantly less efficient than traditional algorithms.

\begin{table}[]\footnotesize
\caption{Rating scale for each solver: Higher values indicate better performance. The highest values are underlined.}
\begin{tabular}{@{}ccccc@{}}
\toprule
\multicolumn{5}{c}{MCP}                                                               \\ \midrule
Method                 & Quality(\%) & Memory(\%)  & Efficiency(\%)  & Robustness(\%) \\ \midrule
Normal Greedy          & 99.73       & \underline{100}   & 0.97            & 98.53          \\
Lazy Greedy (2007)     & \underline{100}   & 37.78       & \underline{100}       & \underline{100}      \\
S2V-QN (2017)          & 87.40       & 18.49       & 0.86            & 9.53           \\
GCOMB (2020)           & 99.11       & 13.27       & 7.43            & 91.58          \\
LeNSE (2022)           & 71.91       & 14.70       & 0.04            & 2.50           \\ \midrule
\multicolumn{5}{c}{IM}                                                                \\ \midrule
Degree Discount (2009) & 89.77       & 96.32       & 82.59           & 8.54           \\
Single Discount (2009) & 89.15       & \underline{96.88} & \underline{84.09}     & 8.33           \\
IMM (2015)             & \underline{99.44} & 60.20     & 12.86           & \underline{100}      \\
OPIM (2018)            & 96.36       & 73.85       & 35.69           & 43.54          \\
Geometric-QN (2020)$*$  & 44.66       & 33.98       & \textless{}0.01 & 5.08           \\
GCOMB (2020)           & 67.71       & 15.15       & 3.06            & 3.41           \\
RL4IM (2021)           & 66.12       & 57.60       & 0.26            & 3.64           \\
LeNSE (2022)           & 78.44       & 29.90       & 0.02            & 5.35           \\ \bottomrule
\end{tabular}

\label{table:rating}
\end{table}

Based on the observed results across a wide range of datasets and settings, a rating scale is summarized in Tab.~\ref{table:rating}. 
Here, the metric \textit{Quality} for method $f$ is defined as the average of $\{c^{(f)}_d/\text{Max}({c_d}) | d \in \text{D}\}$, where $c$ represents the coverage value, and $d$ denotes a selected dataset within the entire set of datasets $D$.  Similarly, the metric \textit{Efficiency} is defined as the average of $\{\frac{\text{Max}({t_d})}{t^{(f)}_d} | d \in \text{D}\}$, where $t$ signifies the runtime. 
Furthermore, we delineate the metric \textit{Robustness} for method $f$ as the normalized reciprocal standard deviation of its quality. Please note that the rating for {\sf Geometric-QN}, due to its limited scalability,  is not derived from a direct comparison with others across $D$. Instead, it is derived by comparing it specifically with {\sf IMM} on the smaller datasets, as detailed in Kamarthi et al.  ~\cite{kamarthi2020influence}.


%% file: 5-Related.tex
\section{Other Related Work}

\noindent {\textbf{Maximum Coverage and Influence Maximization Problems.}} MCP is a well-known NP-hard problem that has received significant attention. Various variants of MCP, such as the Budgeted Maximum Coverage Problem \cite{kar2016budgeted, piva2019approximations, li2021probability, zhou2022effective} and the Multiple Knapsack Problem \cite{kellerer2004multidimensional}, have been extensively studied, finding applications in facilities location \cite{megiddo1983maximum, chauhan2019maximum}, maximum coverage in streams \cite{saha2009maximum}, social recommendation~\cite{wu2023sstp,li2023personalized} and information retrieval \cite{anagnostopoulos2015stochastic}, etc.


The IM problem~\cite{kempe2003maximizing} has also garnered significant attention. By focusing on the diffusion of influence, IM addresses the concept of coverage from a different perspective, making it a natural and relevant extension of the MCP framework in the context of social networks. Kempe et al.~\cite{kempe2003maximizing} proposed a greedy algorithm that provides a ($1-\frac{1}{e}-\epsilon$) approximation under the Independent Cascade (IC) model. Subsequent studies have focused on improving the efficiency and scalability of influence maximization algorithms. For instance, CELF +\cite{leskovec2007cost} and its improved version CELF++ \cite{goyal2011celf++} significantly reduced the Monte Carlo evaluation times while maintaining a $(1 - \frac{1}{e})$ approximation. Other heuristic algorithms \cite{chen2009efficient,wang2012scalable,jiang2011simulated}, offer more efficient solutions without relying on Monte Carlo simulations, although they may not provide strong theoretical guarantees.
To address the computational challenges of influence maximization, Borgs et al. \cite{borgs2014maximizing} introduced the Reverse Influence Sampling (RIS) technique, which achieves nearly linear time complexity relative to the graph size while providing a ($1-\frac{1}{e}-\epsilon$)-approximation under the IC model. Building upon RIS, Tang et al. proposed TIM/TIM++ \cite{tang2014influence} and IMM \cite{tang2015influence}, which further improved the empirical efficiency. Tang et al. also introduced OPIM \cite{tang2018online}, which focuses on influence maximization in online settings.

\noindent {\textbf{Criticism on Machine Learning-based Heuristics for Combinatorial Optimization.}} The remarkable success of machine learning (ML) in diverse fields such as computer vision \cite{krizhevsky2012imagenet}, natural language processing \cite{vaswani2017attention}, and automatic control \cite{mnih2015human} has prompted the belief that ML can excel in other domains as well. However, recent studies have shed light on the limitations of deep learning (DL), both from a broad perspective and in specific applications. \cite{camilleri2017analysing, chollet2017limitations} have discussed these limitations, providing insights into the challenges faced by DL. Furthermore, \cite{cremer2021deep} has examined expert opinions on the potential and limitations of DL, summarizing a body of work that uncovers the inadequate performance of DL in various applications.
A recent contribution~\cite{angelini2022cracking} proposes a benchmark study that highlights the disparity between DL solvers and a simple greedy algorithm in solving the maximum independent set (MIS) problem. Their findings demonstrate that the greedy algorithm not only provides better quality solutions but also exhibits significantly faster computation, surpassing Deep RL solvers by a factor of $10^4$. This work emphasizes the need to carefully evaluate the performance of DL techniques in specific problem domains, as there may be alternative approaches that outperform DL in terms of both solution quality and computational efficiency.

These discussions and benchmark studies serve as important reminders that while DL has achieved remarkable success in numerous areas, it is not a one-size-fits-all solution. Understanding the limitations and exploring alternative approaches can help researchers and practitioners make informed decisions regarding the most suitable techniques for solving specific problems.

%% file: 6-Con.tex
\section{Potential Directions for Improving Deep-RL Methods}
\label{sec:solutions}

Deep-RL methods currently demonstrate excellence primarily in abnormal scenarios with limited practical optimization value, emphasizing their current limitations. In this section, we briefly outline the challenges and limitations of applying Deep-RL to IM.

\noindent \textbf{Identify similar distribution effectively.} As discussed in \S~\ref{sec:concern}, the upper bound of the worst-case performance for Deep-RL methods on MCP and IM is constrained, i.e., not exceeding $1-\frac{1}{e}$. Furthermore, the effectiveness of these methods in scenarios beyond this worst-case heavily depends on their generalizability. This necessitates a thorough assessment of the alignment between the distributions of the training and test graphs. Evidence from experimental results and analyses detailed in \S~\ref{sec:distribution} indicates that this issue remains unresolved. A pivotal challenge is to identify a few easily computable measures to quantify the learned graph distribution.

\noindent \textbf{Extract high-quality query subspace.} The Deep-RL methods often struggle with inefficiency, primarily due to the vastness of the search space they need to explore. To mitigate this, certain Deep-RL approaches aim to enhance efficiency without compromising solution quality by identifying a subgraph within the entire graph that contains the high-quality solution, effectively reducing the search space.
However, this strategy necessitates a delicate balance between the time consumption in identifying the quality of this subspace and the precision of such identification. 
Methods like {\sf Geometric-QN} and {\sf LeNSE} tend to spend more time extracting this subspace compared to the time saved during the querying process. Conversely, while {\sf GCOMB} improves query efficiency by efficiently pruning the search space, it struggles to consistently maintain solution quality. 
Overall, a substantial challenge persists in effectively extracting a high-quality search subspace. 

\noindent \textbf{Initialization with prior knowledge.} Initializing the state with prior knowledge, as opposed to random initialization, holds the potential to enhance performance. 
{\sf Geometric-QN}, for instance, constructs a subgraph based on random initialization, leading to a high variance in the final performance. Similarly, {\sf LeNSE}, by initializing a subgraph randomly, incurs a time-consuming process for subgraph exploration. This suggests that adopting prior knowledge for initialization could potentially streamline exploration processes and yield more stable performance.

\section{Conclusions}

In this benchmark study, we conducted a comprehensive examination of the effectiveness and efficiency of five recent Deep-RL methods for MCP and IM. The experimental results reveal that, for both MCP and IM, traditional algorithms such as {\sf Lazy Greedy} and {\sf IMM} consistently outperform Deep-RL methods in the majority of cases, if not all. Additionally, we highlight common issues observed in Deep-RL methods for MCP and IM, emphasizing the challenge of predicting the performance of a trained Deep-RL model on a testing graph before actual execution. Finally, we discuss potential directions for enhancing Deep-RL methods in the context of MCP and IM. Our benchmark study sheds light on possible challenges and limitations in current deep reinforcement learning research aimed at solving combinatorial optimization problems.



%% file: 7-Appendix.tex
\newpage
\appendix

\section{Details about Lazy Greedy}
The key idea behind the efficiency of the {\sf Lazy Greedy} algorithm, making it significantly faster than a simple greedy algorithm, lies in its utilization of submodularity—a property that ensures diminishing returns as more nodes are added to the selection. This property implies that the marginal gain from adding a new node to a larger set is less than or equal to the marginal gain from adding it to a subset of that set. Alg.~\ref{alg:lazy_greedy} illustrates the pseudocode of {\sf Lazy Greedy}.

\begin{algorithm}
\caption{Lazy Greedy}
\label{alg:lazy_greedy}
\begin{algorithmic}[1]
\State $G \gets (V, E)$ 
\State $K \gets$ budget for the number of nodes
\State $A \gets \emptyset$ \Comment{Initialize the solution set}
\State $\delta s \gets$ number of nodes covered by $s$, for all $s \in V$
\State $Covered \gets \emptyset$ \Comment{Set of currently covered nodes}

\While{$|A| < K$ and $\exists s \in V \setminus A$}
    \State $s^* \gets \arg\max_{s \in V \setminus A} \delta s$
    \State $A \gets A \cup \{s^*\}$
    \State Update $Covered$ to include nodes covered by $s^*$
    \For{$s \in V \setminus A$}
        \State Recompute $\delta s$ as $|N(s) \setminus Covered|$ \Comment{Lazy update}
    \EndFor
\EndWhile
\State \textbf{return} $A$ 
\end{algorithmic}
\end{algorithm}

\section{Noise predictor in GCOMB}
\label{appendix:noise predictor}
GCOMB runs up to two orders of magnitude faster than the simple greedy algorithm and S2V-DQN in MCP is partly credited with the noise predictor. Concretely, GCOMB reduces the search space with aid of filtering out a large number of noisy nodes by noise predictor. Therefore, GCOMB's performance relies heavily on the performance of the noise predictor. Unfortunately, our result demonstrates that the noise predictor, a linear interpolation method in essence, can not learn how to distinguish noisy nodes well.


\subsection{Training time of Noise Predictor}

The noise predictor training method proposed in \cite{manchanda2020gcomb} isn't consistently feasible. In our tests, the noise predictor didn't generalize well to unseen budgets. Consequently, we followed the author's experimental approach, training a distinct noise predictor for each budget. Tab.~\ref{noise-predictor:time} details the time required to train a noise predictor for each budget in MCP. Remarkably, this process takes thousands of times longer than solving a problem using Lazy Greedy.

\begin{table}[h!]
\caption{Training Time of Noise Predictor}
\begin{tabular}{@{}cccc@{}}
\toprule
Budget & DBLP & Youtube & Live Journal \\ \midrule
20     & 13.2 & 33.2    & 384.4        \\
50     & 18.8 & 45.2    & 515.8        \\
100    & 27.9 & 66.5    & 728.9        \\
150    & 36.8 & 88.5    & 951.0        \\
200    & 46.1 & 109.0   & 1181.5       \\ \bottomrule
\end{tabular}
\label{noise-predictor:time}
\end{table}

\subsection{Proportion of Non-Noisy Nodes}
\label{appendix:number}

Tab.~\ref{noise-predictor:number} shows the proportion of good nodes in the entire graph in MCP. It shows that the proportion is not monotone increasing and even varies dramatically, which accounts for the cases where the runtime of GCOMB fluctuates wildly.

What's worse, the noise predictor is unfeasible in some cases. We can find that GCOMB takes much more time than expected to find the solution on Amazon, Pokec, Flixster and Twitter in the IM problem. Simply because the number of good nodes derived from the noise predictor is larger than the total number of nodes in the entire graph, in this case, GCOMB has to find the solution set from the entire graph, which takes a significant amount of time.

\begin{table}[h!]
\vspace{-1mm}
\caption{Number of Non-Noisy Nodes}
\vspace{-2mm}
\begin{tabular}{@{}cccc@{}}
\toprule
Budget & DBLP    & Youtube & Live Journal \\ \midrule
20     & 0.030\% & 0.002\% & 0.01\%       \\
50     & 0.025\% & 0.010\% & 0.04\%       \\
100    & 0.415\% & 0.013\% & 0.013\%      \\
150    & 0.304\% & 0.020\% & 0.022\%      \\
200    & 0.327\% & 0.049\% & 0.061\%      \\ \bottomrule
\end{tabular}
  \label{noise-predictor:number}
\vspace{-3mm}
\end{table}

\section{Improve LeNSE Efficiency}
\label{appendix:lense}
The original implementation of LeNSE is slow performance, requiring several days to generate a training dataset even for small graphs. In our optimization efforts, we have significantly improved the implementation, reducing the preprocessing runtime from days (> 72 hours) to just a few minutes (11.3 minutes), while preserving the underlying logic. In IM problem, our optimizations include utilizing reverse influence sampling techniques for calculating influence spread, and dynamically generating training subgraphs of varying quality. As for MCP, we replace the normal greedy with Lazy Greedy to boost the efficiency during preprocessing and inference.



\section{Discussion on More MCP Variants}

There are some other popular MCP variants, such as Weighted MCP~\cite{nemhauser1978analysis}, Partial Coverage Problem~\cite{gandhi2004approximation}, Stochastic MCP~\cite{goemans2006stochastic}, and Generalized MCP~\cite{cohen2008generalized}. IM is unique due to its relevance in network-driven processes and its ability to simulate the intricate social dynamics where each node's influence is interlinked, presenting a unique strategic complexity.

We acknowledge that each problem poses unique challenges that require tailored Deep-RL solutions. Yet, the inherent similarities across these problems allow for the adaptation of Deep-RL designs with minor modifications. However, the learning difficulty varies significantly with the complexity of each problem, thereby affecting the effectiveness of the application of these methods. Furthermore, when applying Deep-RL methods to other MCP variants, they will unavoidably encounter similar challenges as seen in MCP and IM.


We provide a detailed comparison between IM with other popular MCP variants as follow:

\noindent \textbf{Weighted MCP~\cite{nemhauser1978analysis}}

\begin{itemize}
    \item Description: In Weighted MCP, each element in the universe \(U\) has an associated weight \(w(e)\), and the objective is to maximize the total weight of the elements covered by the selected subsets.
    \item Mathematical Formulation: \[
     \text{Maximize} \sum_{e \in U} w(e) \cdot x_e
     \]
     where \(x_e = 1\) if element \(e\) is covered by any of the selected sets, and 0 otherwise.
    \item Comparison with IM: While Weighted MCP introduces complexity through element weights, it operates under deterministic conditions where each set’s coverage is predefined. IM, conversely, deals with probabilistic influences where the effectiveness of each node is uncertain and subject to network dynamics. This uncertainty adds a layer of strategic decision-making absent in the more predictable scenarios of Weighted MCP.
\end{itemize}

\noindent \textbf{Partial Coverage Problem~\cite{gandhi2004approximation}:}
\begin{itemize}
    \item Description: The objective is to achieve a specified level of coverage \(C\), which might not be the maximum, using a limited number of sets.
    \item Mathematical Formulation:
    \begin{align*}
        \text{Maximize}\quad & \sum_{e \in U} x_e \\
        \text{s.t.} \quad & \sum_{e \in U} x_e \geq C
    \end{align*}
     
     \item Comparison with IM: The Partial Coverage Problem, where the goal is to achieve a predefined level of coverage, also operates under deterministic or less complex probabilistic settings. IM extends this by requiring coverage under highly dynamic and uncertain conditions, reflecting the unpredictable nature of social networks and the influence processes within them.

\end{itemize}

\noindent \textbf{Stochastic MCP~\cite{goemans2006stochastic}:}
\begin{itemize}
    \item Description: Coverage of each set is uncertain, defined by probabilistic measures, with each set \(S_i\) having a probability \(p_{i,e}\) of covering element \(e\).
    \item Mathematical Formulation:
     \[
     \text{Maximize} \sum_{e \in U} \left(1 - \prod_{i:e \in S_i}(1 - p_{i,e})\right)
     \]
     \item Comparison with IM: While both Stochastic MCP and IM involve dealing with probabilities, IM introduces the concept of network effects. In the context of IM, especially under models like the Independent Cascade and Linear Threshold models, the influence of one node depends on the states of others, known as network effects. This implies that the probability of one node influencing another is contingent upon the collective state of its neighbors, thus introducing strategic complexity.
\end{itemize}

\noindent \textbf{Generalized MCP~\cite{cohen2008generalized}:}
\begin{itemize}
    \item Description: The Generalized Maximum Coverage (GMC) Problem involves selecting subsets (bins) and elements to maximize profit under a collective budget constraint, considering individual weights and profits for elements across various bins.
    \item Mathematical Formulation:
     \begin{align*}
    \text{Maximize} \quad & \sum_{e \in \eta} P(f(e), e) \\
    \text{s.t.} \quad & \sum_{b \in \beta} W(b) + \sum_{e \in \eta} W(f(e), e) \leq L
    \end{align*}

    \item Comparison with IM: Unlike IM, which is focused on the spread of influence in a network with inherent stochastic dependencies between nodes, GMC is concerned with maximizing profit by efficiently allocating resources across multiple bins under a budget constraint. While IM utilizes network dynamics where the outcome (influence spread) is influenced by the probabilistic interactions between nodes, GMC deals with a deterministic selection process aimed at optimizing resource usage without the direct influence of network effects. The strategic complexity in IM arises from the need to manage and leverage these network effects, which does not directly apply in the context of GMC.
\end{itemize}

Based on these insights, IM was selected for our study due to its relevance in network-driven processes and its ability to simulate the intricate social dynamics where each node's influence is interlinked, presenting a unique strategic complexity. This focus on IM not only fills a significant gap in our understanding of influence mechanisms within extensive networks but also lays a foundation for applying these strategies to other complex coverage problems.

\section{Additional Results}
In the MCP experiment, we present supplementary results for twelve datasets. Additionally, in the IM experiment, we display further results for six datasets under the edge weight models WC, TV, and CONST, respectively. Due to scalability issues, LeNSE and RL4IM crash on the dataset Orkut.

\begin{figure*}[h]
  \centering
  \includegraphics[width=\linewidth]{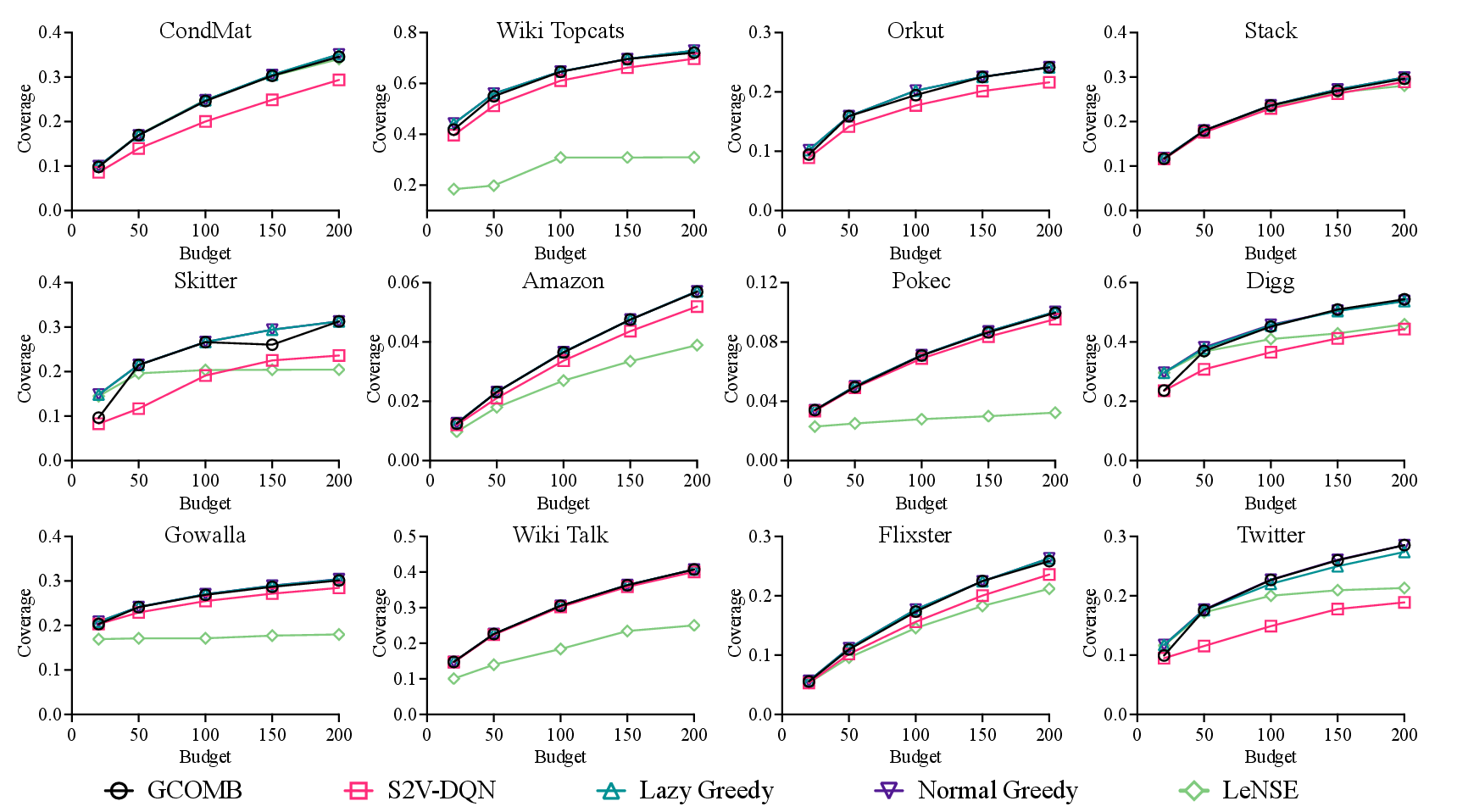}
  \caption{Additional results for MCP: coverage curve}
  \label{fig:all_MCP_coverage}
\end{figure*}

\begin{figure*}[h]
  \centering
  \includegraphics[width=\linewidth]{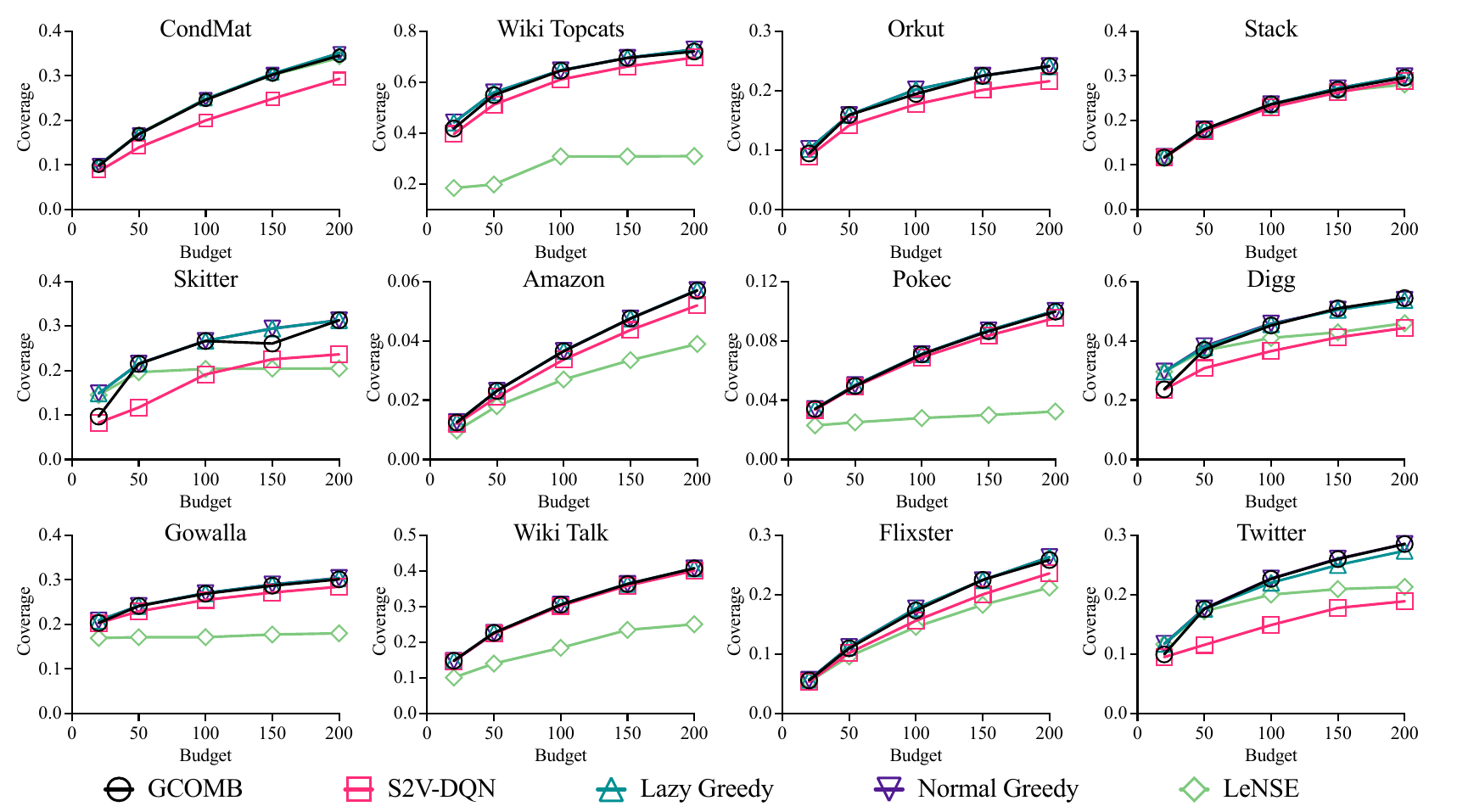}
  \caption{Additional results for MCP: runtime curve}
  \label{fig:all_MCP_runtime}
\end{figure*}

\begin{figure*}[h]
  \centering
  \includegraphics[width=\linewidth]{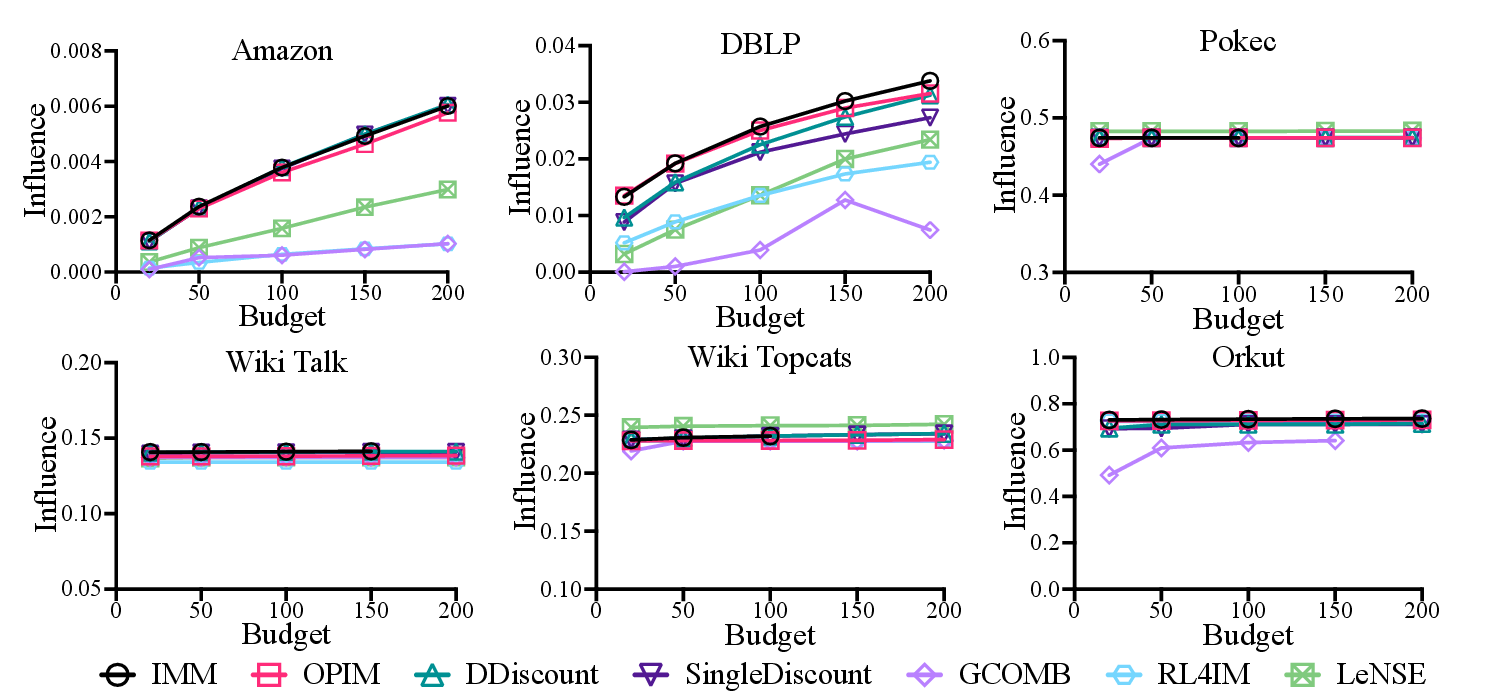}
  \caption{Additional results for IM: Influence curve under CONST}
  \label{fig:const_coverage}
\end{figure*}

\begin{figure*}[h]
  \centering
  \includegraphics[width=\linewidth]{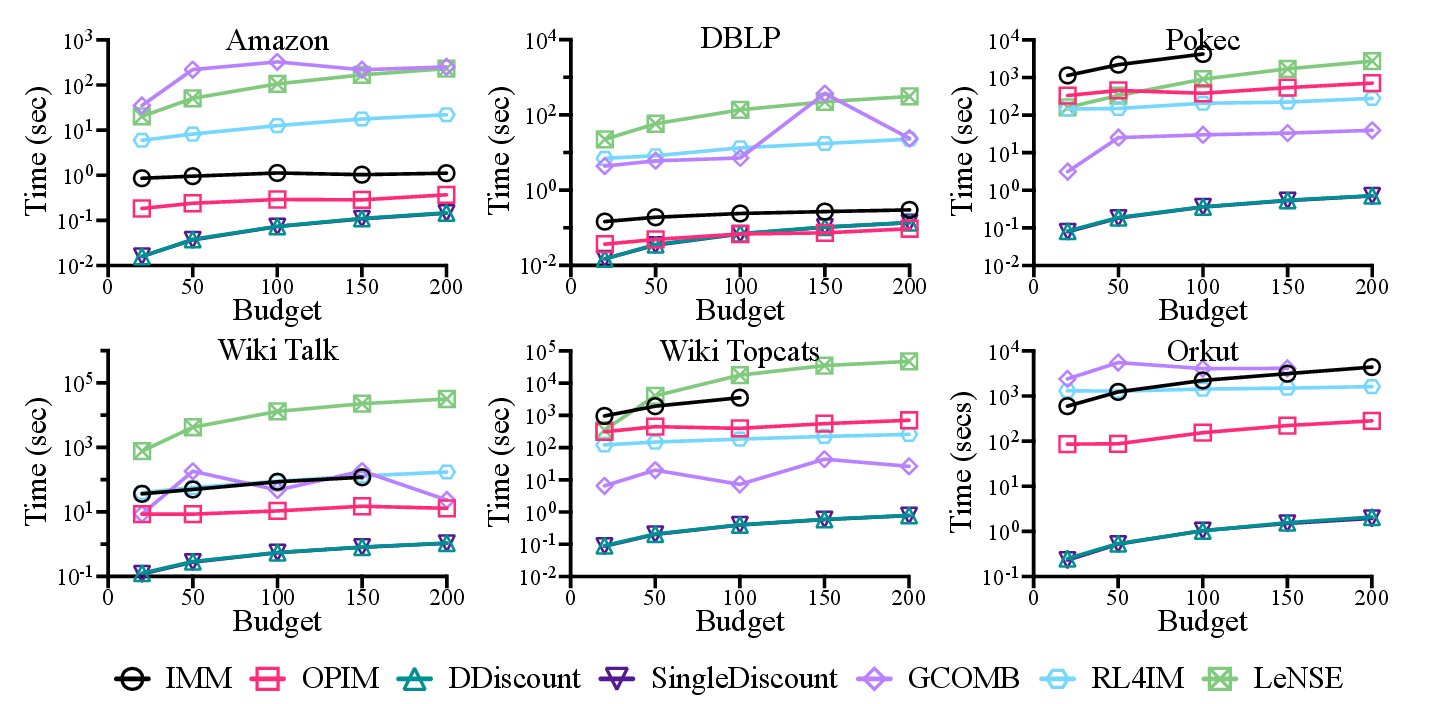}
  \caption{Additional results for IM: Runtime curve under CONST}
  \label{fig:const_efficiency}
\end{figure*}

\begin{figure*}[h]
  \centering
  \includegraphics[width=\linewidth]{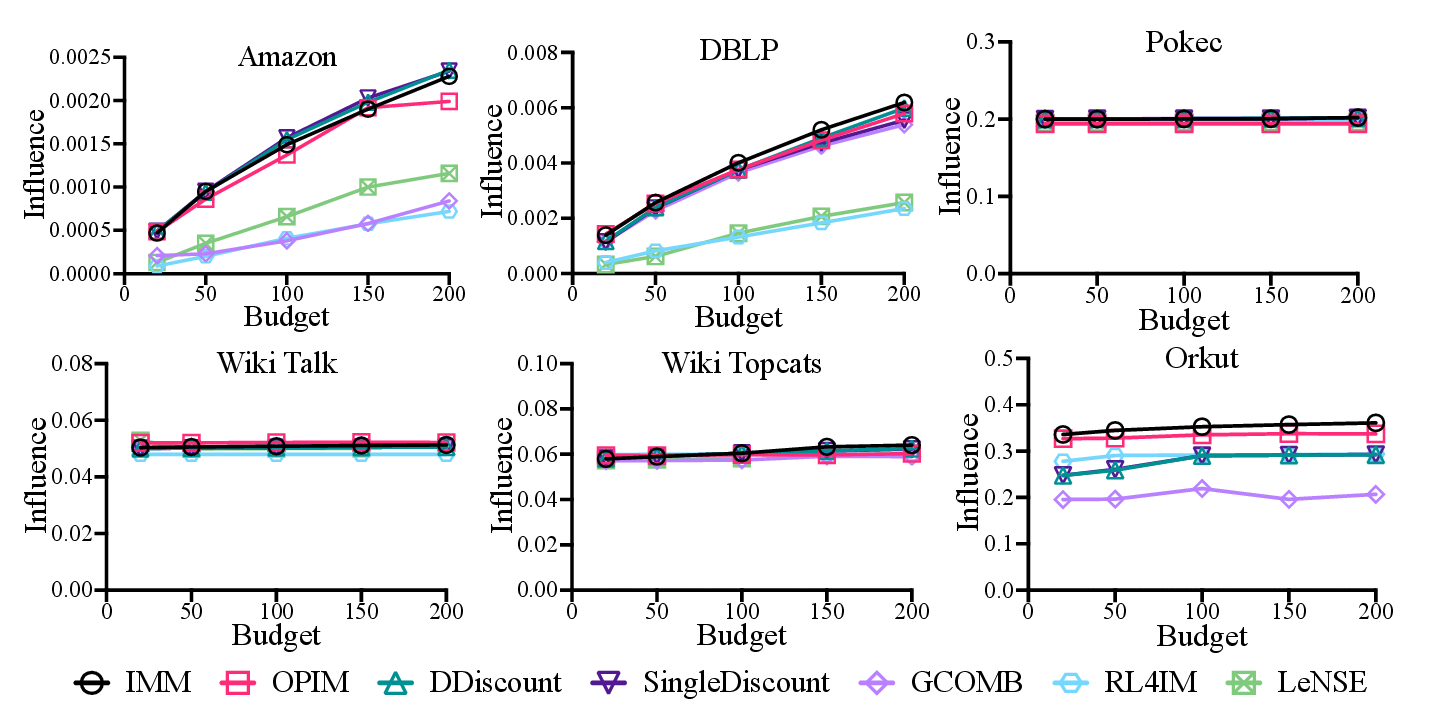}
  \caption{Additional results for IM: Influence curve under TV}
  \label{fig:tv_coverage}
\end{figure*}

\begin{figure*}[h]
  \centering
  \includegraphics[width=\linewidth]{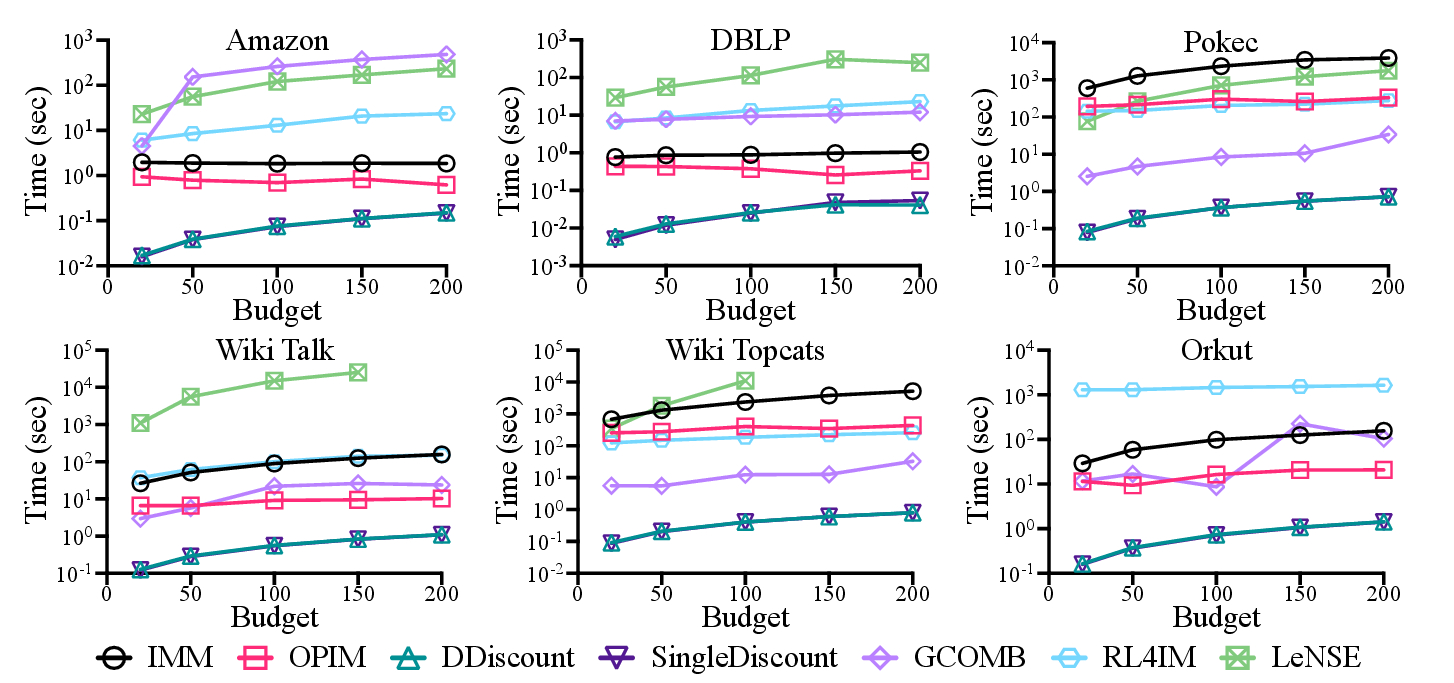}
  \caption{Additional results for IM: Runtime curve under TV}
  \label{fig:tv_efficiency}
\end{figure*}

\begin{figure*}[h]
  \centering
  \includegraphics[width=\linewidth]{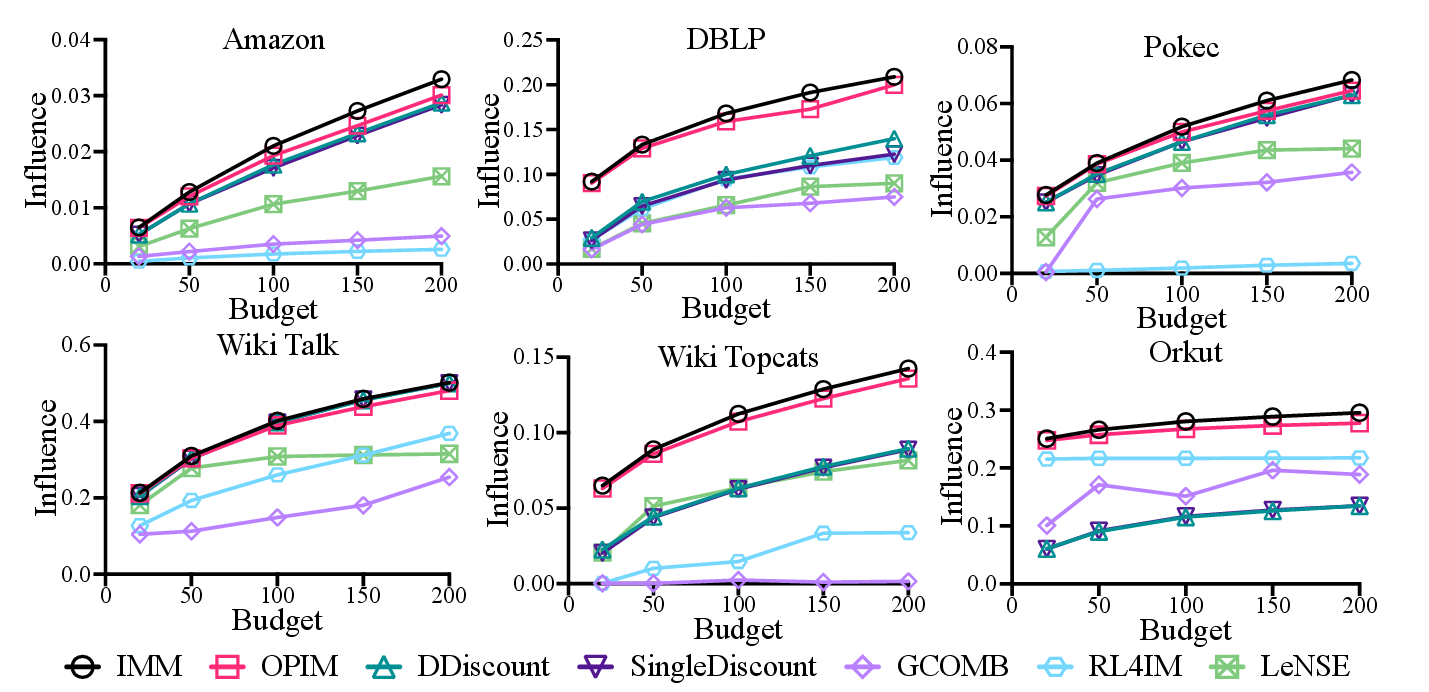}
  \caption{Additional results for IM: Influence curve under WC}
  \label{fig:wc_coverage}
\end{figure*}

\begin{figure*}[h]
  \centering
  \includegraphics[width=\linewidth]{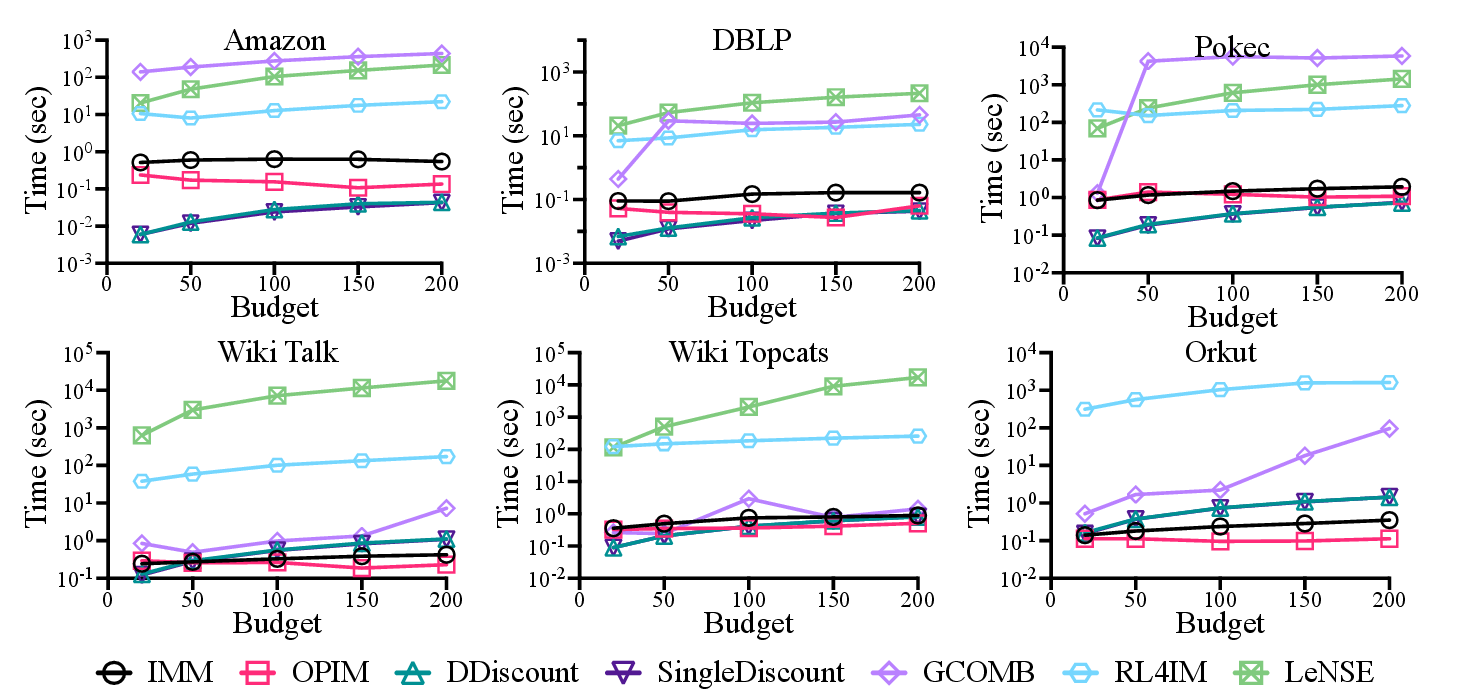}
  \caption{Additional results for IM: Runtime curve under WC}
  \label{fig:wc_efficiency}
\end{figure*}